\documentclass[sigconf]{acmart}
\usepackage{graphicx}
\usepackage{amsmath}
\usepackage{booktabs}
\usepackage{multirow}
\usepackage{amsmath,bm}
\usepackage{float}
\usepackage{makecell}
\usepackage{balance}
\usepackage[capitalize]{cleveref}
\crefname{section}{Sec.}{Secs.}
\Crefname{section}{Section}{Sections}
\Crefname{table}{Table}{Tables}
\crefname{table}{Tab.}{Tabs.}
\newcommand{\ori}{\textcolor{black}}
\newcommand{\hut}{\textcolor{black}}
\newcommand{\hutnew}{\textcolor{black}}
\AtBeginDocument{%
  }

\setcopyright{acmcopyright}
\copyrightyear{2018}
\acmYear{2018}
\acmDOI{XXXXXXX.XXXXXXX}

\copyrightyear{2023}
\acmYear{2023}
\setcopyright{acmlicensed}\acmConference[MM '23]{Proceedings of the 31st ACM International Conference on Multimedia}{October 29-November 3, 2023}{Ottawa, ON, Canada}
\acmBooktitle{Proceedings of the 31st ACM International Conference on Multimedia (MM '23), October 29-November 3, 2023, Ottawa, ON, Canada}
\acmPrice{15.00}
\acmDOI{10.1145/3581783.3611766}
\acmISBN{979-8-4007-0108-5/23/10}
\acmSubmissionID{340}

\begin{document}

\title{Stroke-based Neural Painting and Stylization with  Dynamically Predicted Painting Region}


\author{Teng Hu}
\orcid{0009-0008-1247-5931}
\affiliation{%
  \institution{Shanghai Jiao Tong University}
  \city{Shanghai}
  \country{China}
}
\email{hu-teng@sjtu.edu.cn}

\author{Ran Yi}
\authornote{Corresponding author.}
\affiliation{%
  \institution{Shanghai Jiao Tong University}
  \city{Shanghai}
  \country{China}
  }
\email{ranyi@sjtu.edu.cn}

\author{Haokun Zhu}
\affiliation{%
  \institution{Shanghai Jiao Tong University}
  \city{Shanghai}
  \country{China}
  }
\email{zhuhaokun@sjtu.edu.cn}

\author{Liang Liu}
\affiliation{%
  \institution{Youtu Lab, Tencent}
  \city{Shanghai}
  \country{China}
}
\email{melpancake@gmail.com}

\author{Jinlong Peng}
\affiliation{%
  \institution{Youtu Lab, Tencent}
  \city{Shanghai}
  \country{China}
  }
\email{jeromepeng@tencent.com}

\author{Yabiao Wang}
\affiliation{%
  \institution{Youtu Lab, Tencent}
  \institution{Zhejiang University}
  \city{Shanghai}
  \country{China}
  }
\email{caseywang@tencent.com}

\author{Chengjie Wang}
\affiliation{%
\institution{Shanghai Jiao Tong University}
  \institution{Youtu Lab, Tencent}
  \city{Shanghai}
  \country{China}
}
\email{jasoncjwang@tencent.com}

\author{Lizhuang Ma}
\affiliation{%
  \institution{Shanghai Jiao Tong University}
  \city{Shanghai}
  \country{China}
}
\email{ma-lz@cs.sjtu.edu.cn}
\renewcommand{\shortauthors}{Teng Hu et al.}

\begin{abstract}

  Stroke-based rendering aims to recreate an image with a set of strokes. Most existing methods render complex images using a\hut{n} uniform-block-dividing strategy, which leads to boundary inconsistency artifacts. 
To solve the problem, we propose \textbf{Compositional Neural Painter}, a novel stroke-based rendering framework which dynamically predicts the next painting region based on the current canvas, instead of dividing the image plane uniformly into painting regions.
We start from an empty canvas and divide the painting process into several steps. At each step, a compositor network trained with a phasic RL strategy first predicts the next painting region, then a painter network trained with a WGAN discriminator predicts stroke parameters, and a stroke renderer paints the strokes onto the painting region of the current canvas.
Moreover, we extend our method to stroke-based style transfer with a novel differentiable distance transform loss, which helps preserve the structure of the input image during stroke-based stylization. 
Extensive experiments show our model outperforms the existing models in both stroke-based neural painting and stroke-based stylization. Code is available at: \href{https://github.com/sjtuplayer/Compositional_Neural_Painter}{https://github.com/sjtuplayer/Compositional\_Neural\_Painter}.
\end{abstract}

% <concept_id>10010147.10010178.10010224</concept_id>
% <concept_desc>Computing methodologies~Computer vision</concept_desc>
% \ccsdesc[500]{Computing methodologies~Computer vision}

\keywords{stroke-based rendering, phasic RL strategy, distance transform }


\maketitle

\ccsdesc[500]{Computing methodologies~Computer vision}

\section{Introduction}
\label{sec:intro}
\begin{figure}[t]
\centering
\includegraphics[width=0.45\textwidth]{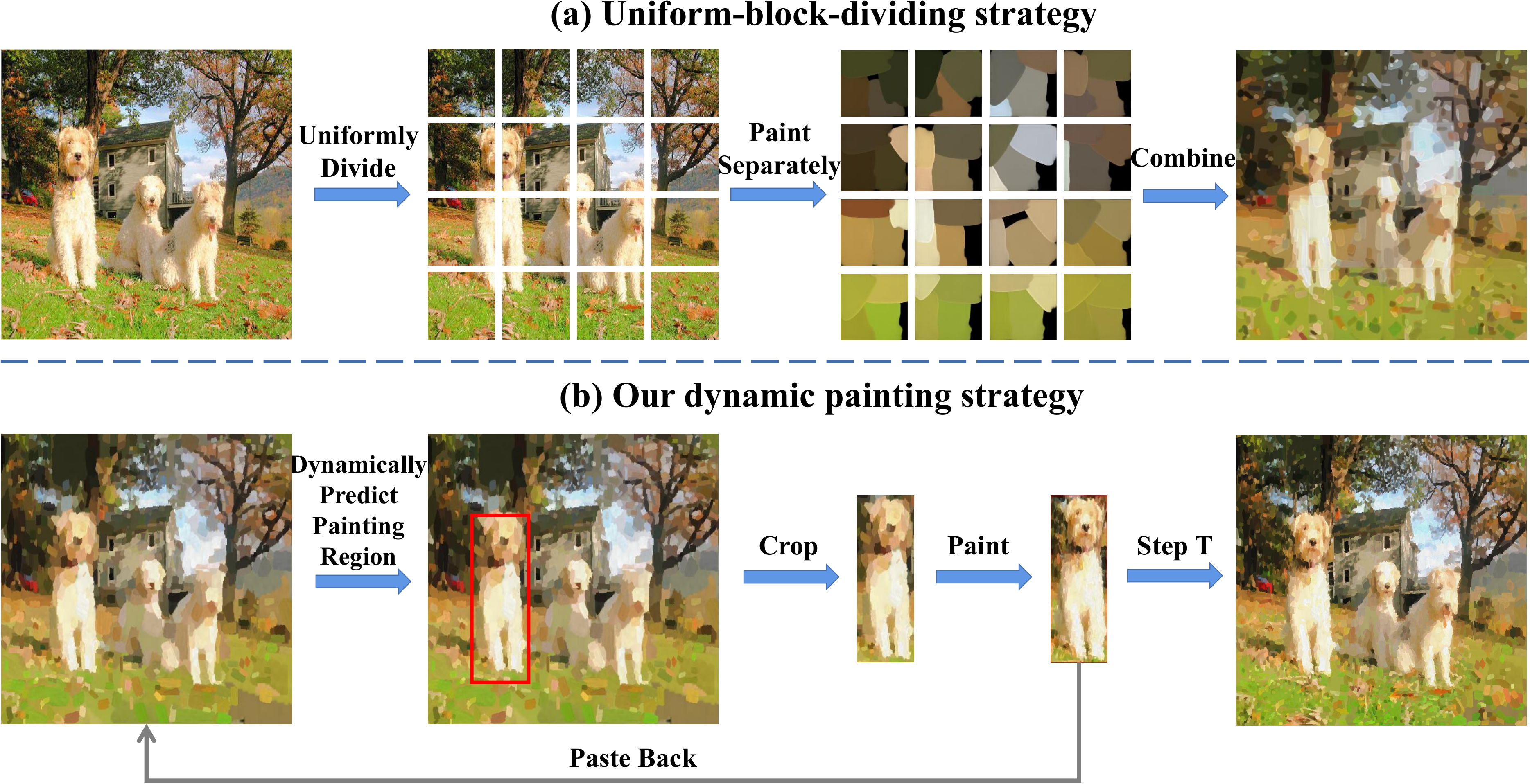}
\vspace{-0.05in}
\caption{(a) The painting process of existing models~\cite{huang2019learning,liu2021paint,zou2021stylized} which uniformly divide the image plane into $k\times k$ blocks and paint each block independently, resulting in boundary inconsistency artifacts between each two adjacent blocks; (b) The painting process of our model: we dynamically predict the next painting region (red box) and paint it with our painter network, which avoids the boundary inconsistency artifacts and reconstructs more details in the target image.}
\label{fig:paiting process}
\vspace{-0.15in}
\end{figure}
% Painting has been an important way for humans to record information and express emotions since ancient times. Creating a well-drawn painting in the past generally needs expert painters to spend a long time on it, making it difficult for ordinary people to participate in such artistic creation. In recent year, as graphics processing technology evolved, especially the development of deep-learning technology, many works are dedicated in image generation and helping people to create artistic images from real photos or texts. 
Stroke-based rendering (SBR) aims to recreate an image with a set of brushstrokes. Different from mainstream generation models based on VAE ~\cite{kingma2013auto}, GANs ~\cite{goodfellow2014generative} and Diffusion model ~\cite{ho2020denoising}, which generate images using pixels as basic elements, SBR uses brushstrokes as basic elements and decomposes the painting process into a stroke sequence. By painting the strokes sequentially onto the canvas, SBR can better imitate the painting process of human artists. %painters do which are more like the real artistic works.

\begin{figure*}[t]
\centering
\includegraphics[width=0.9\textwidth]{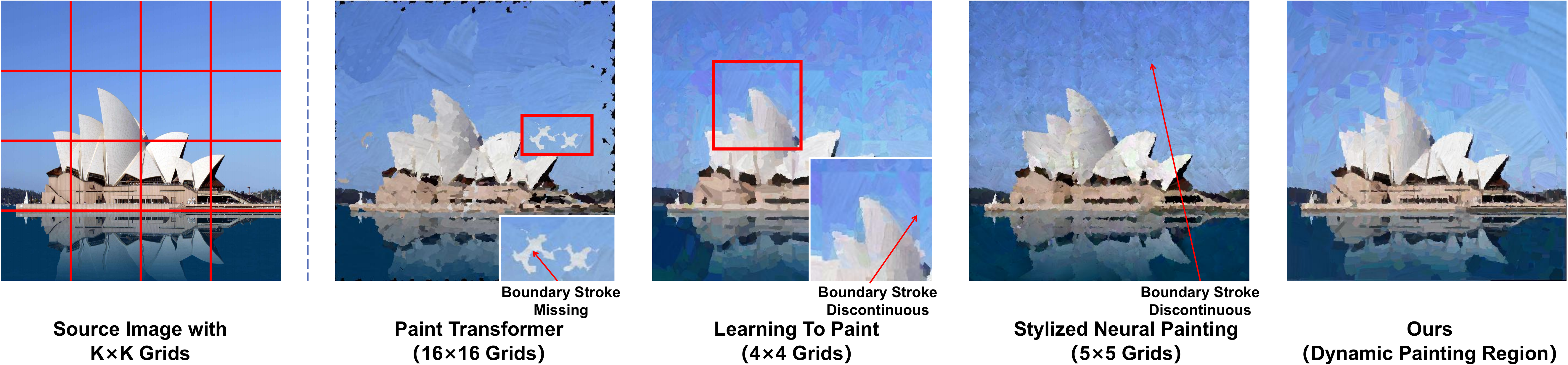}
\vspace{-0.05in}
\caption{\hut{The boundary inconsistency artifacts of the existing methods with 3,000 strokes: Paint Transformer ~\cite{liu2021paint}, Learning To Paint ~\cite{huang2019learning} and Stylized Neural Painting ~\cite{zou2021stylized}. All the methods suffer from the discontinuous strokes at the junction of the adjacent blocks. %which makes the paintings unreal. 
Please zoom in for details.}}
\label{fig:boundary inconsistency}
\vspace{-0.15in}
\end{figure*}

\hut{Traditional} SBR methods ~\cite{hertzmann1998painterly,litwinowicz1997processing,turk1996image,haeberli1990paint,teece19983d,ganin2018synthesizing,zhou2018learning,tong2022im2oil} rely on \hut{non-deep learning} methods, e.g., greedy searching and \hut{heuristic optimization} with low efficiency.  
% \hutnew{Among them, Im2Oil~\cite{tong2022im2oil} is the latest and most representative one, featuring excellent painting effects.}
Recent \hut{deep learning-based} methods can be classified into three classes: RL-based methods ~\cite{huang2019learning,singh2021combining,singh2022intelli}, DL-based methods ~\cite{liu2021paint} and optimization-based methods ~\cite{zou2021stylized,kotovenko2021rethinking}. 
All the existing methods can only predict a limited number of strokes for an input image \hut{block}, since more strokes require much more training \hut{or optimization} resources. 
But the images in real world (e.g., images in ImageNet ~\cite{deng2009Imagenet}) are usually too complex to reconstruct with limited strokes. 
To render more details, existing methods ~\cite{huang2019learning,liu2021paint,zou2021stylized} uniformly partition the image plane into $k\times k$ blocks and predict strokes for each block independently. 
However, this uniform-block-dividing strategy (Fig.~\ref{fig:paiting process}(a)) suffers from the following weaknesses:
(1) it is only used in their testing stage, which makes their testing setting inconsistent with their training setting;
(2) it leads to boundary inconsistency artifacts (Fig. \ref{fig:boundary inconsistency}), i.e., since each block is rendered separately, the blindness to adjacent blocks leads to inconsistent strokes on the two sides of the block boundary.%(stroke discontinuity or hiatus in .
% Recent works ~\cite{huang2019learning,zou2021stylized} employ reinforcement learning (RL) to train a SBR painter network. Paint Transformer~\cite{liu2021paint} abandon RL to train the painter, but it needs much more strokes than the Rl methods to achieve the same reconstruction quality. Besides , there is also some optimization-based method  

% Although some of the models ~\cite{huang2019learning,liu2021paint,zou2021stylized} can build a quite better image decomposition for real images, they all suffer from a serious problem which we call it the boundary inconsistency. In fact, None of these models can render a good result with their training settings. In training stage, they train the model to render an image directly.
% But in the testing stage, they divide the target image into $k\times k$ blocks first and render each block respectively. Although the result in each block is not so accurate, by concatenating them together, the final output can be much more similar to the target image in a global sight. This uniform-block-dividing strategy improve the models performance amazingly, but dealing with the block images separately leads to the confusion at the block boundary, shown as stroke discontinuity or hiatus in Figure \ref{fig:boundary inconsistency}. We summarize the confusions at the boundary as boundary inconsistency.

Inspired by the real painting process, where artists usually decide the painting region first, and then draw the objects in the corresponding region, %\ori{This drawing strategy} helps human artists paint more accurately and vividly.
%This model architecture comes from the real painting process which can be summarized as the where-to-paint stage and the how-to-paint stage.
we propose Compositional Neural Painter, a novel stroke-based rendering framework with a painting scheme that first predicts {\it ``where to paint''} and then decides {\it ``what to paint''}. 
% \hut{Compared to Intelli-Paint~\cite{singh2022intelli} which employs sliding attention window to guide the painting process in the local foreground object region and strongly relies on the object detection method, our model can predict the  painting regions in a global view and is free from reliance on any object detection models. This allows our model to be more robust and effective in handling scenarios with multiple or no clearly defined foreground objects}
\hutnew{Compared to Intelli-Paint~\cite{singh2022intelli} which employs sliding attention window to guide the painting process in the local foreground object region and strongly relies on the object detection method, our model can predict the  painting regions in a global view and is free from reliance on any object detection models. This allows our model to be more robust and effective in handling scenarios with multiple or no clearly defined foreground objects}
%上一段讲清楚动态预测画框并对比，下一段开头衔接好

\hut{In detail, our model is composed of two parts: a compositor network predicting {\it ``where to paint''} and a painter network decides {\it ``what to paint''.}}
The compositor network is proposed to dynamically predict the next painting region based on the current canvas (Fig.~\ref{fig:paiting process}(b)), instead of dividing the image plane uniformly into painting regions.
We start from an empty canvas and decompose the painting process into several steps.
At each step, the compositor network first predicts the next painting region, a painter network then predicts the stroke parameters, and a stroke renderer paints the strokes onto the painting region of the canvas.
Specifically, the compositor network is trained with a RL strategy with phasic reward function; and the painter network is a CNN-based model trained with a WGAN discriminator to penalize always painting similar strokes, which often happens without using RL strategy.
Furthermore, we extend our method to stroke-based stylization %to improve the performance of stylization in SBR, we propose a new SBR-based stylization method 
with a novel differentiable distance transform loss, which helps preserve the structure of the input image during stroke-based stylization. %when doing stroke-based stylization.

The contributions of our work are three-fold:
\begin{itemize}

\item We propose Compositional Neural Painter, a novel stroke-based rendering \hut{model} which dynamically predicts the next painting region based on the current canvas. This dynamic rendering strategy solves the boundary inconsistency artifacts caused by the uniformly divided painting regions in existing methods and achieves good \hut{painting performance}. %reconstruction accuracy.
\item We propose a compositor network trained with a phasic RL strategy to predict the next painting region, a painter network trained with a WGAN discriminator to \hut{forecast } the stroke parameters \hut{in the predicted painting region}, and a neural renderer for stroke rendering.
% \item
% We propose a new RL-free training method to train the painter network which outperforms the existing models in both accuracy and efficiency.
% \item
% We propose the  RL-free training method to train the painter network which can achieve an even better performance than the RL-based state-of-the-art methods.
\item We extend our method to stroke-based style transfer \hut{task} with a novel differentiable distance transform loss, which helps preserve the structure of the input image during stroke-based stylization.
\end{itemize}

\section{Related Works}
{\bf Stroke-based rendering (SBR)}. Stroke-based image rendering (SBR) aims at recreating a target image with a set of brushstrokes. 
Different from the general image synthesising models (e.g., VAE ~\cite{kingma2013auto}, GANs ~\cite{goodfellow2014generative} and Diffusion Models ~\cite{ho2020denoising}) which generate images using pixels as basic elements, \hut{SBR methods emulates the real painting process of human artists, employing brushstrokes as the fundamental unit to paint stroke-by-stroke, thereby enhancing the fidelity of the painted images in terms of local texture and brushstroke details to that of real artistic works.}
% SBR uses brushstrokes as the basic painting elements,.
%It is dedicated to mimicking the techniques used by human painters, employing strokes as the fundamental building block of the image and putting the final image together through a succession of stroke sequences. 
The traditional SBR algorithms either employ greedy search ~\cite{hertzmann1998painterly,litwinowicz1997processing,tong2022im2oil}, devise heuristic optimization ~\cite{turk1996image}, or require user inputs ~\cite{haeberli1990paint,teece19983d} to find the position and other characteristics of each stroke. \hutnew{Among them, Im2Oil~\cite{tong2022im2oil} is the latest method, which incorporates adaptive sampling and greedy search based on probability density maps, resulting in superior painting results.} 
With the development of deep learning in recent years, many SBR methods based on neural networks have been proposed. 
Early works ~\cite{graves2013generating,ha2017neural} employed recurrent neural networks (RNN) to decompose the image into brushstrokes. However, the demand for manual annotation limits their application. ~\citet{ganin2018synthesizing} and ~\citet{zhou2018learning} introduced reinforcement learning (RL) to synthesize stroke sequences, but can only render simple images like sketches.
\begin{figure*}[t]
\centering
\includegraphics[width=0.9\textwidth]{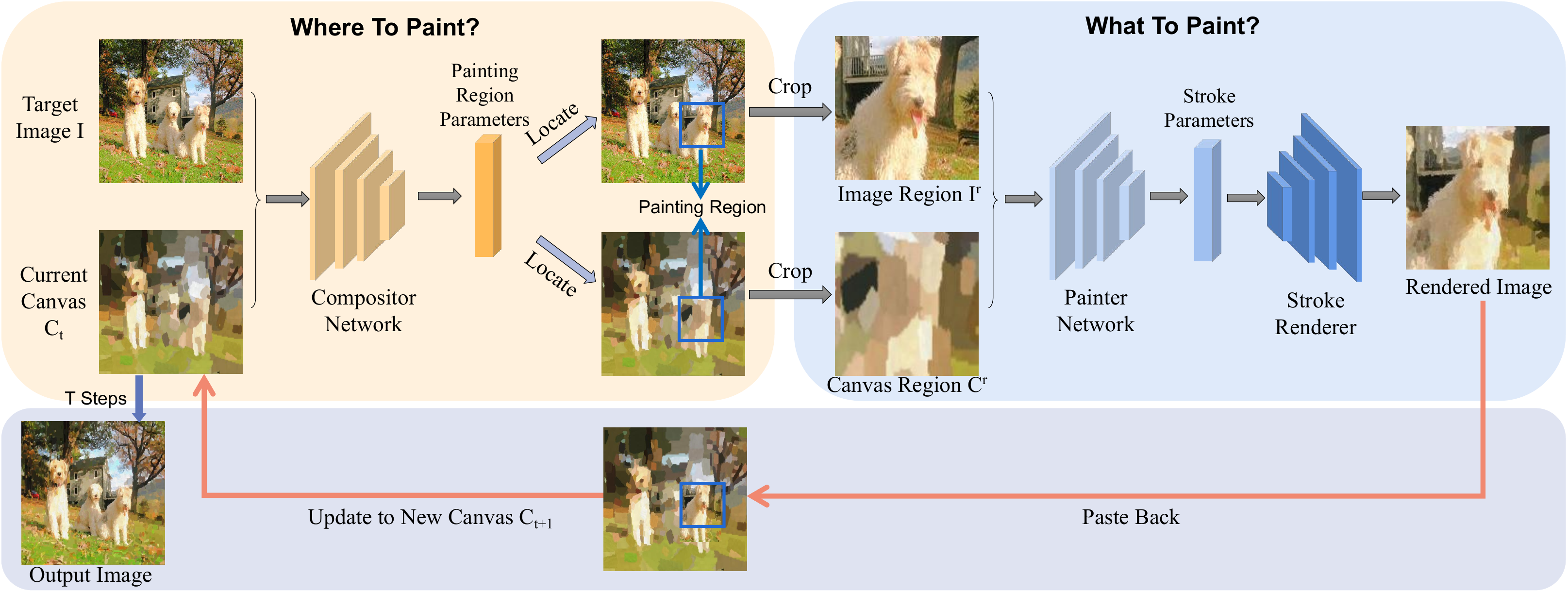}
\vspace{-0.1in}
\caption{\hut{Model framework of our Compositional Neural Painter}: Our model consists of a compositor network, a painter network and a stroke renderer. We start from an empty canvas and decompose the painting process into T steps. At each step $t$, the compositor network first predicts the painting region parameters $r_t$, then the painter network estimates the strokes parameters $s_t$ and the stroke renderer paints the strokes constructed from $s_t$ onto the painting region $r$ of the current canvas $C_t$. After T steps, our model output the painting constructed by a set of strokes.}
\label{fig:model framework}
\vspace{-0.1in}
\end{figure*}

%{\bf Neural SBR for real images.} 
To paint a complex real-world images using stroke-based rendering, ~\citet{huang2019learning} employed a more complicated RL model DDPG ~\cite{lillicrap2015continuous} with a WGAN ~\cite{arjovsky2017wasserstein} reward. %which can do a more complex image decomposition. 
To improve the reconstruction ability, ~\citet{singh2021combining} and \hut{~\citet{singh2022intelli}} introduced semantic guidance into the RL model, making it concentrate more on the main object in the image.
\hut{Specifically, Intelli-Paint~\cite{singh2022intelli} predicts both attention window and stroke parameters in one network to better imitate human painting process, but its heavy reliance on object detection and the limited RL training efficiency restricts its ability in painting complex images and predicting enough strokes for fine-grained details. To solve the low-training efficiency of the RL-based methods,}
% However, the RL-based methods often suffer from a long training time and an unstable agent. 
~\citet{liu2021paint} then proposed a RL-free model Paint Transformer, which accelerates the training stage and achieves better training stability.
Besides the learning-based methods, some optimization-based methods ~\cite{zou2021stylized,kotovenko2021rethinking} decomposed the images into a series of parameterized strokes through iterative optimization process. But they suffer from a long optimization time for each image. 
Although the above methods can achieve a relatively good result in real image rendering, they either suffer from the boundary inconsistency artifacts, or lack the ability to render a more complex image. 
\ori{In contrast, we employ a compositor network to dynamically predict the next paint region and then paint by a painter network, which not only solves the boundary inconsistency artifacts, but also achieves a better reconstruction quality.} 
\hutnew{Moreover, in contrast to Intelli-Paint\cite{singh2022intelli}, which utilizes sliding attention windows to guide the painting process in the foreground object region and heavily relies on object detection, our two-stage painting approach has the ability to paint significantly more intricate details without the need of object detection.}
 
{\bf Stroke-based style transfer.} 
Style transfer aims at transferring the style from a style image to a content image.
%To capture style from the target image, ~\citet{gatys2016image} leverages the image features extracted by a VGG network~\cite{simonyan2014very} to explicitly describe the style information. 
Previous style transfer methods ~\cite{gatys2016image,huang2017arbitrary,cao2018carigans,liu2021adaattn,song2021agilegan} confined the stylization process in the pixel domain.
stroke-based style transfer methods ~\cite{zou2021stylized,kotovenko2021rethinking} stylized images using brushstrokes as the basic element, optimizing stroke parameters instead of pixels. 
%However, since the original stylization method ~\cite{gatys2016image} is used at pixel level, the existing SRB models can not maintain the structure of the source image well. So, the stylization in SBR still needs some improvement.
% In the past, redrawing an image in a particular style required a professional artist to spend a lot of time on creating. In recent years, with the development of deep learning technology, there comes numerous researches working on style transfer.
However, the existing stroke-based style transfer methods cannot preserve of the structure of the input image well.
\ori{Different from these methods, we design a new stroke-based stylization framework with a novel differentiable distance transform loss, which can preserve the structure of the input image during stroke-based stylization.}

\section{Method}
\subsection{Overview}
Stroke-based rendering aims at recreating an image using strokes as basic painting elements. Existing methods ~\cite{huang2019learning,liu2021paint,zou2021stylized,singh2021combining,singh2021combining,singh2022intelli} predict a limited number of strokes  for an input image \hut{block}, since more strokes require much more training  \hut{or optimization} resources. 
But the images in real world are usually too complex to reconstruct with limited strokes.  To render more details, existing methods ~\cite{huang2019learning,liu2021paint,zou2021stylized} uniformly partition the image plane into $k\times k$ blocks, and predict stroke parameters for each block independently. However, this uniform-block-dividing strategy in the test stage leads to boundary inconsistency artifacts:
since the strokes of each image block are predicted separately, 
the blindness to the adjacent blocks leads to inconsistent strokes on the two sides of the block boundary. Fig.~\ref{fig:boundary inconsistency} shows some examples of the boundary inconsistency artifacts, including stroke-discontinuous and stroke-missing \hut{problem}. \hut{Besides, the semantic-based methods~\cite{singh2021combining,singh2022intelli} abandon uniform-block-dividing strategy and concentrate on the foregrounds by painting more strokes. } \hutnew{But their heavy reliance on object detection  and the low RL-training efficiency restrict their ability in painting complex images with multiple or no clearly defined objects.}

To solve these problems, we propose {\it Compositional Neural Painter}, a novel stroke-based rendering model that dynamically predicts the next painting region based on the current canvas.  
By dynamically deciding the painting region instead of uniformly partition, our method can better reconstruct details and solve the boundary inconsistency problems.
%learns from the way human artists paint and can draw paintings stroke by stroke in a coarse-to-fine manner. 
%Our model takes an image $I$ as input and output the rendered image $I_r$ composed of brushstrokes.
Our Compositional Neural Painter consists of three modules: 1) a compositor network that takes a target image $I$ and a canvas $C$ as inputs and predicts the next painting region $r$ 
% that indicates where should be painted next
; 2) a painter network that takes the cropped target image $I^r$ and the cropped canvas $C^r$ according to region $r$ as inputs, and predicts the stroke parameters $s$; 3) a stroke renderer that renders the strokes of parameters $s$ back into the current canvas.

We start from an empty canvas $C_0$ and decompose the painting process into $T$ steps (Fig.~\ref{fig:model framework}).
At each step $t$, the compositor network first predicts the painting region $r_t$, then the painter network estimates the strokes parameters $s_t$ for $N$ strokes, and the stroke renderer paints the strokes constructed from $s_t$ onto the painting region $r$ of the current canvas $C_t$.
% In each step $t$, the compositor network and painter network input the source image and current canvas alter
% inputs the current canvas $C_t$ and source image $I$ together and outputs the position $p_t$ where should be further drawn by the painter network. Then, the painter network takes in the partial canvas $C_t^p$ and partial image $I^p$ cropped from the original canvas $C_t$ and image $I$ as inputs and predicts the strokes parameters $s$ which will be rendered on the partial canvas $C_t^p$ by the renderer.As last, the partial canvas $C_t^p$ will be put back to the whole canvas $C_t$, which outputs the new canvas $C_{t+1}$.
After $T$ steps, we %get a sequence of painting regions and stroke parameters $\{(r_t,s_t), 0\le t\le T-1\}$. 
%By drawing strokes $\{s_t^1,s_t^2,\cdots,s_t^N\}$ on each position $p_t$, we can 
get a final rendered image $I_r=C_{T}$ made up of sequentially painted brushstrokes, which resembles the target image $I$.

%We'll introduce each block in our model one by one.

\subsection{Compositor Network: Where To Paint?}

In real painting process, human artists usually decide where to paint first based on the current canvas, and then paint the strokes in the corresponding painting region, instead of painting from top left to right down. %The two-stage drawing strategy helps human artists paint more accurately and vividly. In the where-to-paint stage, human artists analyse the  painting in a global sight and choose a part to draw next. Then, in the drawing stage, artists only need to focus on the drawn part and ignore the influence of the unrelated parts which can improve the drawing accuracy and efficiency. 
Inspired by this, we design the painting process of a set of strokes as human artists do: first predicting ``where to paint'' and then deciding ``what to paint''.
To solve the first question, we design a compositor network, which dynamically predicts the next painting region based on the current canvas, instead of uniformly partitioning the image plane into painting regions.
The compositor network has a global sight of the painting and guides the whole model to paint in a coarse-to-fine manner.
 
%  In real painting process, human artists usually draw a painting in two steps: a coarse-to-fine composition step and a drawing step. In the composition stage, human artists arrange the placement and drawing order of visual elements in a coarse-to-fine manner. They tend to draw the overall outline first, and then gradually enrich the details and textures of the image. In the drawing stage, since the composition has been made, artists only need to focus on the drawn part which can improve the drawing accuracy and efficiency. Inspired by this, we divide the painting process as human artists do into two stages : where to paint and how to paint. To solve the first question, we design a compositor network who has a global sight of the painting and guides the whole model to paint in a coarse-to-fine manner.

% Our painter network can be regarded as a professional painter who is well-skilled in local painting. So, we design a compositor network who has a global sight of the painting and guides the painter network to paint from a coarse-to-fine manner.

{\bf Model framework.} At each step $t$, the compositor network takes a target image $I$ and a current canvas $C_t$ as inputs \hutnew{to} predict the painting region $r_t$ indicating where should be painted next. The painting region is a rectangular region denoted by $r_t=(x,y,w,h)$, where $(x,y)$ refers to the upper-left endpoint coordinate, and $(w,h)$ denote the width and height. After predicting the painting region $r_t$, the painter network and stroke renderer will predict the stroke parameters for this region and render the strokes onto the current canvas to get a new canvas $C_{t+1}$.

% $$Reward=
% \begin{cases}
% 0& \quad \quad \quad \quad  \quad\, r\ge ^-5\times10^-3\\
% 0&  2\times 10^{-3}\le r\le5\times10^{-3}\\
% 1& \quad \quad \quad \quad  \quad\, r\le 2\times10^{-3}
% \end{cases}$$
{\bf Training strategy.}
\hut{Previous works train the neural painting  networks based on self-supervision~\cite{huang2019learning,singh2021combining,liu2021paint,singh2022intelli}  (minimizing the distance between the rendered image and input image). However, since}
 our painting process needs to crop the canvas according to the predicted painting region, %the digital images are represented by pixels, 
the predicted parameters $(x,y,w,h)$ need to be rounded to the nearest integers. %to indicate the  rectangular to-be-drawn part. 
Due to the non-differentiable rounding operation, deep learning strategies  based on back propagation are not applicable to train the compositor network. %can not be trained directly. 
Therefore, we introduce a Reinforcement Learning (RL) framework DDPG ~\cite{lillicrap2015continuous} to train the compositor network\footnote{Note that we first train stroke renderer and painter network, and then train the compositor network with the other two networks fixed.}. %It is worth emphasizing that our compositor network only have a small legal searching space spanned by the parameters $(x,y,w,h)$, which do not need a long time to train.

The original reward function $R$ used in DDPG-based RL methods ~\cite{huang2019learning,singh2021combining} is formulated as:
\begin{align}
    R_{ori}=\frac{\|I-C_t\|_2}{\|I\|_2}-\frac{\|I-C_{t+1}\|_2}{\|I\|_2}~.
    \label{eq:original reward}
\end{align}

The original reward function aims to minimize the $\mathcal{L}_2$ distance between the canvas $C_{t+1}$ and the target image $I$. 
However, as $C_t$ gradually converges to $I$, the pixel loss between \hutnew{them becomes extremely small}, making it difficult for the critic network to capture the subtle variations.
To solve this problem, we design a phasic reward mechanism based on the pixel loss between the drawn canvas $C_{t+1}$ and the target image $I$: when the pixel loss is \hutnew{below} a threshold, we enlarge the reward with a non-linear mapping. \ori{Denote $d=\frac{\|I-C_{t+1}\|_2}{\|I\|_2}$}, the phasic reward function is formulated as:
% \begin{align}
%     R_{phasic}=f(\frac{\|I-C_t\|_2}{\|I\|_2}-\frac{\|I-C_{t+1}\|_2}{\|I\|_2}),
% \end{align}
\begin{equation}
\begin{aligned}
R_{phasic}&=\beta R_{ori}, &\quad
\beta = \left\{
	\begin{aligned}
	&1, &\quad d>0.005\\
	&f(1-d)~, &\quad d\le 0.005\\
	\end{aligned}
	\right
	.
\end{aligned}
\end{equation}
where $f(x)=\frac{1}{\alpha}\ln{\frac{1+x}{1-x}}$ is the \ori{inverse sigmoid function} and $\alpha$ is a constant.  \hut{It's worth noting that in practical experiments, we add a small $\epsilon=10^{-6}$ to the denominator of $f(x)$ to avoid the situation \hutnew{where $x$ is extremely close to $1$ (i.e., $d$ approaches $0$)} .  }

Compared to the original reward function (Eq.(\ref{eq:original reward})), our new reward function enlarges the small reward in the later training steps, which helps our model to reconstruct more fine-grained details and textures in the target image.
% Experiments shown in the Sec\ref{sec:expeiment on reward function} demonstrate that this design is very effective.
%%%%effective?

\subsection{Painter Network: What To Paint?}
The painter network aims to reconstruct an input image $I$ using a sequence of strokes. %which can reconstruct the contents of $I$. 
%This can be summarized as an stroke set prediction problem whose goal is to minimize the distance between the input and output images. 
Most of the existing methods ~\cite{ganin2018synthesizing,zhou2018learning,huang2019learning,singh2021combining} employ RL to train their models. %but suffers from a long training time and a low training efficiency. 
For example, Learning To Paint ~\cite{huang2019learning} needs 5 additional neural networks to help train one painter network. %, which leads to a long training time.  
A recent work Paint Transformer ~\cite{liu2021paint} abandons RL and accelerate the training stage, but it needs much more strokes than the RL models to achieve the same reconstruction accuracy.
We find that a CNN based painter network achieves better reconstruction accuracy than existing methods when working with a WGAN discriminator, and requires a much simpler training procedure.

% In the RL models ~\cite{ganin2018synthesizing,huang2019learning,singh2021combining}, they employ a specially designed WGAN discriminator to calculate the reward. Different from the ordinary discriminator which takes the real images $x$ as the \hutnew{real} samples and the generated images $\overline{x}$ as the \hutnew{fake} samples, their discriminator take the the concatenation of two $x$ as the \hutnew{real} samples and the concatenation of $x$ and $\overline{x}$ as the \hutnew{fake} samples. This makes the judgement of the \hutnew{real} sample too easy by simply checking whether the concatenated images are the same. According to the GANs theory~\cite{goodfellow2014generative}, a too strong discriminator gives a poor guidance to the generator (can be regarded as the painter network here). In this section, we rethink the function of adversarial learning in SBR in a new angle and propose a new RL-free method to train our painter network with different WGAN framework.

{\bf Model framework.} %We propose a CNN-based painter model which have both the high training efficiency and good reconstruction ability at the same time. 
%Our painter network draw the paintings from an empty canvas in T steps. 
At each step $t$, the painter takes the target image $I$ and the current canvas $C_t$ as inputs, and outputs $N$ brushstrokes' parameters $s_t=\{s^{(1)}_t,s^{(2)}_t,\cdots,s^{(N)}_t\}$. Then, the stroke renderer $R$ is employed to render these stroke parameters into a stroke image $I_s=R(s_t)$ and a binary mask $M_s$, which are then pasted into the current canvas to get the new canvas $C_{t+1}$:
\begin{align}
    C_{t+1}=I_s\odot M_s +C_t\odot (1-M_s)~,
\end{align}
where $\odot$ is the element-wise multiplication.
After $T$ steps, we get the final output image $I_r=C_T$ which reconstructs the content in the target image $I$.

% \begin{figure}[t]
% \centering
% \includegraphics[width=0.47\textwidth]{figures/pixel-vs-w loss2.pdf}
% \caption{Comparison of $\mathcal{L}_{pixel}$, Learning to Paint ~\cite{huang2019learning} and Ours ($\mathcal{L}_{pixel}+\mathcal{L}_{adv}$). The painter trained with pixel loss alone can not reconstruct the target image. 
% With $\mathcal{L}_{adv}$\hut{,} our painter achieves a better reconstruction ability than the RL method \hut{(Learning to Paint)} ~\cite{huang2019learning}.}
% \label{fig:pixel vs wgan}
% \vspace{-0.1in}
% \end{figure}

{\bf Training strategy.} We train the painter network with a WGAN discriminator. %We update our painter network in each painting step.
%For an input image pair ($I$, $C_t$) and the output image $C_{t+1}$, 
We first minimize the $\mathcal{L}_2$ distance between the new canvas and the target image:
\begin{align}
    \mathcal{L}_{pixel}=\|I-C_{t+1}\|_2~.
\end{align}

However, only optimizing the pixel loss $\mathcal{L}_{pixel}$ leads to poor reconstruction accuracy ~\cite{ganin2018synthesizing}. %leads to a model collapse in the training process.
Simply put, the model trained with the pixel loss only tends to repeatedly generate similar coarse strokes \hut{(refer to ablation study in Sec. \ref{ssec:ablation})} and fails to paint image details.

To penalize the painter network from painting similar strokes, we introduce a WGAN discriminator into the training process. We design a discriminator network $D$ which takes the generated images as the \hutnew{fake} samples and aims to penalize generating similar strokes. Since $D$ has seen the images generated in the previous iterations, once the painter paints the similar strokes again, $D$ will easily discriminate the new canvas as fake and penalize it. In this way, the discriminator constantly pushes the painter to explore different strokes and well reconstruct the target image. %finally find the best solution.
Our painter network can be regarded as a generator, and the training process employs a WGAN-GP~\cite{gulrajani2017improved} loss function:
\begin{align}
    \mathcal{L}_{adv}=D(C_{t+1})-D(x)-\lambda(\|\nabla_{\hat{x}}D(\hat{x})\|_2-1)^2~,
\end{align}
where $x$ is a real image, $C_{t+1}$ is a generated canvas and $\hat{x}$ is a interpolation between $C_{t+1}$ and $x$. It' worth noting that since the discriminator only aims to penalize the strokes seen before, the \hutnew{real} sample $x$ can be any images even a random noise (refer to experiments in Sec. \ref{ssec:ablation}).
%值得注意的是，RL也有一个WGAN，输入形式是xx，目标是xx，再讲问题，用不到我们这里，最后说我们的不同
% \hut{Moreover, different from  Learning to paint whose WGAN discriminator takes a concatenation of two real images as \hutnew{real} sample and a concatenation of one real image and one generated image as \hutnew{fake} sample to guide the training process, our approach adheres to the standard GAN discriminator setting where we use real images as \hutnew{real} samples and generated images as \hutnew{fake} samples to penalize the creation of repetitive strokes and to facilitate model training.}
\hut{\hutnew{Moreover, our discriminator is different from that in} Learning To Paint, \hutnew{which} uses the concatenation of two identical real images as \hutnew{real} sample, and the concatenation of one real image and the corresponding painted image as \hutnew{fake} sample. This makes the discriminator only need to determine whether the two concatenated images are the same \hutnew{to discriminate between the real and fake samples.} \hutnew{Therefore, it's not necessary for the discriminator to remember and penalize the painted images it has seen, which may make it difficult to effectively penalize the previously seen painted images and result in the generation of duplicate brushstrokes.} In contrast, our discriminator only focuses on penalizing the creation of repetitive strokes which facilitates model training.}

% where the \hutnew{real} samples $x$ can be any images (except the \hutnew{fake} samples $C_{t+1}$) even a random noise and $\hat{x}$ is a interpolation between $C_{t+1}$ and $x$.
Then the total loss function is:
\begin{align}
    L_{total}= \mathcal{L}_{pixel}+\gamma\mathcal{L}_{adv}~,
\end{align}
% In summary, the total optimizing process can be summarized as a min-max problem:
% \begin{align}
%     P^*,D^*=\min\limits_{P}\max_{D} \mathcal{L}_{pixel}+\gamma \mathcal{L}_{adv}
% \end{align}
where $\gamma=\lambda\frac{\| \mathcal{L}_{pixel}\|}{\|\mathcal{L}_{adv}\|}$ is an adaptive regularization factor \hut{which balances the two loss functions}, and $\lambda$ is a constant.
\subsection{Stroke Renderer}
% \begin{figure}[t]
% \centering
% \includegraphics[width=0.48\textwidth]{figures/stroke image.pdf}
% \caption{Transform the basic stroke into the binary stroke mask $M_s$ and colored stroke image $I_s$ through the stroke parameters $s=\{x,y,w,h,\theta,r,b,g\}$}
% \label{fig:stroke image}
% \vspace{-0.1in}
% \end{figure}
The stroke renderer aims to render a stroke image $I_s$ and a binary stroke mask $M_s$ based on the stroke parameters.
In our model, we use a real brushstroke \hut{(oil brushstroke~\cite{zou2021stylized})} as the basic stroke and transform it into strokes with different properties according to the parameters.
\ori{The strokes parameters are $s=\{x,y,w,h,\theta,r,g,b\}$, where $(x,y)$ indicate the coordinate of the stroke center, $w,h$ are the width and height of the stroke, $\theta$ is the rotation angle and $(r,g,b)$ is the RGB color of the stroke.}
% The transformation process is shown in Figure \ref{fig:stroke image}.
% The basic stroke is first transformed into the binary stroke mask $M_s$ by the shape parameters $\{x,y,w,h,\theta \}$. Then 

We need to construct a differentiable renderer to render the stroke image and mask from the stroke parameters. %从stroke参数，我们可以用放射变换来得到目标图像
However, the \hut{affine transformation} methods \hut{that transform images through translation, rotation, and scaling parameters} in Computer Graphics (CG) are usually non-differentiable. Following the existing works ~\cite{huang2019learning,zou2021stylized,liu2021paint}, we employ a neural network to render the strokes. 
Specifically, the neural renderer takes the \ori{stroke} parameters as inputs and output \hut{the binary stroke mask $M_s$ and the stroke image $I_s$ where $I_s=(M_s*r,M_s*g,M_s*b)$ }. %Then, the stroke image $I_s$ is calculated by:
%\begin{align}
%    I_s=(M_s*r,M_s*g,M_s*b)
%\end{align}
%\ori{With the stroke parameter s, we first apply affine transformation to the basic stroke image to get the transformed stroke mask $\hat{M}_s$. Then, taking $\hat{M}_s$ as the target, we train our renderer by minimizing the loss function:}
Taking the mask $\hat{M}_s$ rendered from \hut{affine transformation} as ground-truth, the neural renderer is trained using the following loss function:
\begin{align}
    \mathcal{L}_{renderer}=\| \hat{M}_s-M_s \|_2~.
    \label{eq:renderer loss}
\end{align}

\hut{We follow Learning To Paint~\cite{huang2019learning} to train our renderer. Furthermore, } our stroke renderer can also output an additional binary edge map $E_s$ of the stroke \ori{(with the same training method as $M_s$)}, which will be used in the edge-promoting SBR style transfer (Sec.~\ref{ssec:stylization}).

\subsection{Stroke-based Style Transfer}
\label{ssec:stylization}
Style transfer aims at transferring the style form one image to another while maintaining the content. Previous style transfer methods ~\cite{gatys2016image,huang2017arbitrary,liu2021adaattn} confined the stylization process \hutnew{at pixel level}. Some stroke-based style transfer methods ~\cite{zou2021stylized,kotovenko2021rethinking} stylized images using brushstrokes as the basic element, optimizing stroke parameters instead of pixels. However, the existing stroke-based style transfer methods cannot preserve of the structure of the input image well.

To solve this problem, we extend our method to stroke-based style transfer with a novel differentiable distance transform loss which can help preserve the structure of the input image. For style transfer task, our stroke renderer additionally renders the stroke parameters into an edge map $E_s$. By pushing the renderred edge map to fit the edge map of the input image, we can get an edge-promoting stylization output which can keep the structure of the input image as much as possible.

We approximate the distance transform matrix using a differentiable operation, \ori{which calculates the minimum distance between a pixel to each edge pixel (white in edge map) in its surrounding $K^2$ region $\mathcal{N}$.}
We first build a DT kernel $D=(d)_{K\times K}$ for the surrounding $K^2$ region of a pixel $(i,j)$,
where each element $d$ is the Euclidean distance to the pixel $(i,j)$.
Then, we approximate the distance transform matrix of edge map $E$ by:
\begin{equation}
\begin{aligned}
DT(E)_{(i,j)}=\min_{(k,l)\in \mathcal{N}(i,j)} [&E_{(k,l)}\cdot D_{(k-i,l-j)}\\+&(1-E_{(k,l)})\cdot d_{max}]~,\\
\end{aligned}
\label{eq:DT matrix}
\end{equation}
where $d_{max}$ is the maximum distance \ori{in the kernel}. 
 
%The minimum operation in Eq.(\ref{eq:DT matrix}) is non-differentiable.
We further introduce a continuous function to replace the minimum operation \hut{as follows}:
\begin{align}
      min(d_1,d_2,\cdots d_n)=\lim_{\lambda\rightarrow 0} \sum_i \frac{e^{\frac{-d_i}{\lambda}}}{\sum_j e^{{\frac{-d_j}{\lambda}}}}\cdot d_i~,
\end{align}
\hut{where} we set $\lambda=0.3$ in experiments.

After getting the differentiable distance matrix, we calculate the distance transform loss between the two edge maps $E_s$ and $E$ \hut{by}:
\begin{align}
     \mathcal{L}_{DT}(E_s,E)=\mathbb{E}_{(i,j)} [DT(E_s)_{(i,j)} E_{(i,j)}]~.
\end{align}

In stroke-based stylization, given an input image $I_s$ and a style image $I_t$, we first render a stroke-based image $I_r$ from the input image $I_s$ using our Compositional Neural Painter. 
With the stroke parameters $S=\{s_1,s_2,\cdots,s_n\}$ from $I_r$, we can render an edge image $E_s$ by the stroke renderer. 
Denote the binary edge map of the input image as $E_{gt}$, we optimize the stroke parameters $S$ by the following loss function:
\begin{equation}
\begin{aligned}
    S^*=\min\limits_S [\lambda_{style}\mathcal{L}_{style}+\lambda_{con} \mathcal{L}_{con}+\lambda_{DT} \mathcal{L}_{DT}(E_s,E_{gt})]~,
\end{aligned}
\end{equation}
where $\mathcal{L}_{style}$, $\mathcal{L}_{con}$ are style and content loss from ~\cite{gatys2016image}.
\begin{figure*}[!htbp]
\centering
\includegraphics[width=0.97\textwidth]{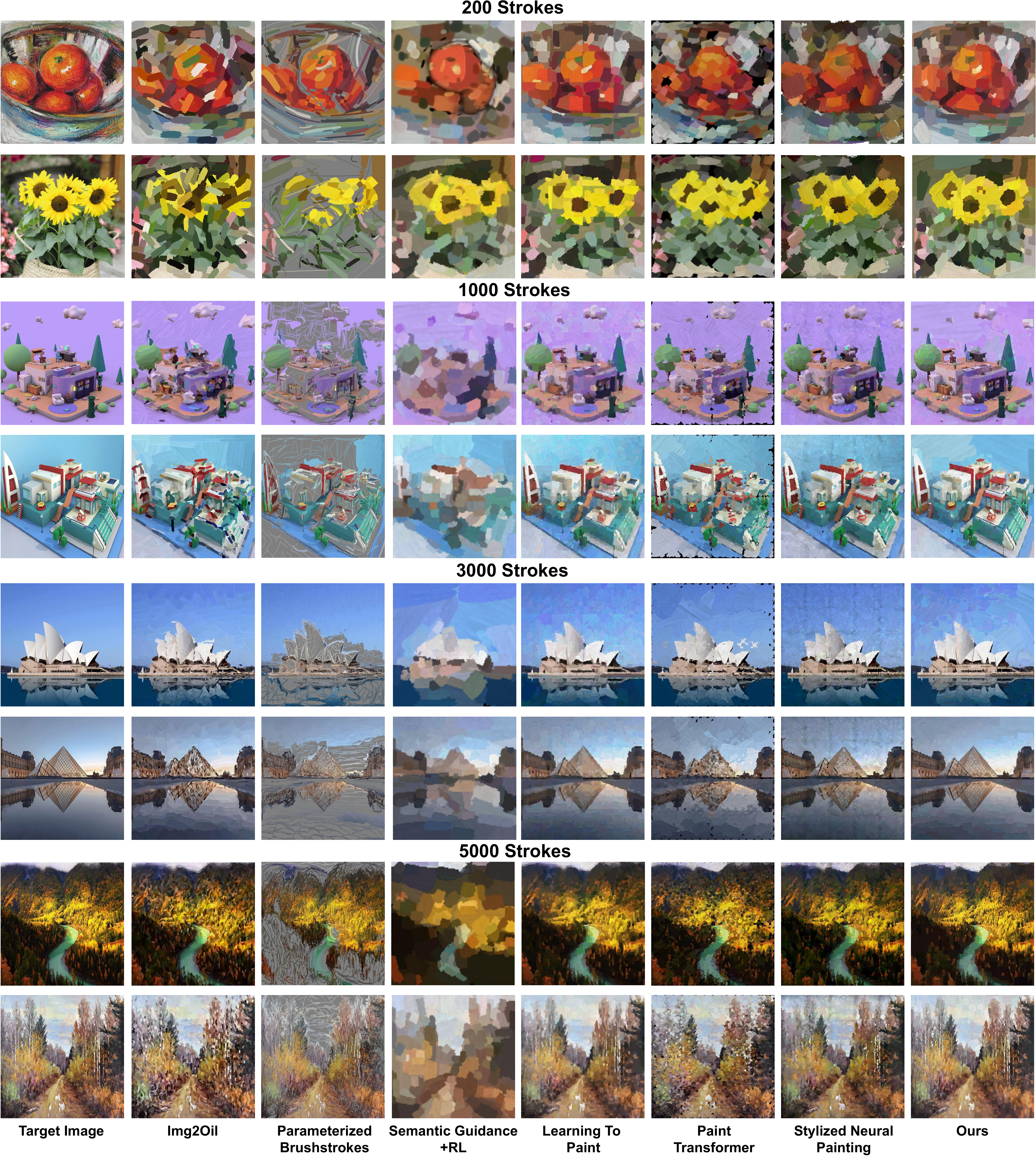}
\caption{The comparison between our model and the state-of-the-art neural painting methods under 200, 1,000, \hut{3,000} and 5,000  strokes. 
Neither Parameterized Brushstrokes ~\cite{kotovenko2021rethinking} nor Semantic+RL ~\cite{singh2021combining}  can reconstruct the images well. 
Learning To Paint ~\cite{huang2019learning}, Paint Transformer ~\cite{liu2021paint} and Stylized Neural Painting ~\cite{zou2021stylized} all suffer from the boundary inconsistency artifacts \hut{and Im2Oil ~\cite{tong2022im2oil} has a better visual results, but lacks some details especially when the stroke number is limited (e.g., 200 strokes)} \hut{In contrast,} our result not only solves \hut{the boundary inconsistency problem}, but also generates a better results with more details and textures. Please zoom in for more details
% (Note that Stylizaed Neural Painting can render at most 2,400 strokes due to excessive demand for memory, we report its results with 2,400 strokes in the 5,000 strokes setting).
}
\label{fig:image to painting}
\end{figure*}

\section{Experiments}

\subsection{Datasets and Settings}
{\bf Datsets.} We conduct experiments mainly on two datasets: CelebA-HQ\cite{karras2018progressive} and ImageNet ~\cite{deng2009Imagenet}. %, which are implemented in both Learning To Paint ~\cite{huang2019learning} and Semantic Guidance+RL ~\cite{singh2021combining}. 
We train and test all the compared models on the two datasets separately. For each dataset, we randomly pick out 1,000 images for testing and the remaining for training.

{\bf Evaluation metrics.} We evaluate the SBR results using the following \hut{three} metrics:

(1) {\bf $\bm{\mathcal{L}_2}$ Distance} calculates the mean $\mathcal{L_2}$ distance between the rendered images and the target images at pixel level. A lower $\mathcal{L}_2$ distance indicates a better image reconstruction quality.

(2) {\bf PSNR:} Peak Signal to Noise Ratio (PSNR) is one of the most commonly and widely used image quality evaluation metrics. A higher PSNR indicates a better image reconstruction quality.

(3) {\bf $\bm{\mathcal{L}_{PIPS}}$ ~\cite{zhang2018unreasonable}} is a perceptual metric to measure the similarity between two images. A lower $\mathcal{L}_{PIPS}$ denotes a higher similarity between the rendered and target images.

% {\bf FID: }Fréchet Inception Distance~\cite{heusel2017gans} measures the distance between the CNN features extracted from the real data and generated data based on their mean values and covariance. A lower FID score indicates a better generation quality.

% {\bf KID: }Kernel Inception Distance~\cite{binkowski2018demystifying} calculates the squared Maximum Mean Discrepancy (MMD) between the CNN features of the real and generated data . KID is more dependable than FID when the test data is insufficient because of its unbiased estimator. Lower KID denotes less of a difference between real and generated data.
\subsection{\ori{Training Details}}
\ori{The training process consists of the three steps:}

(1) \ori{{\bf Train stroke renderer.} We first train our stroke renderer with synthesised data. In detail, we randomly sample stroke parameters $s$ and transform the basic stroke to get the target stroke mask \hut{$\hat{M}_s$}. With the data pair \hut{$(s,\hat{M}_s)$}, we train our renderer by minimizing Eq.(\ref{eq:renderer loss}) for 1M iterations with batch size 32.}

(2) \ori{{\bf Train painter network.}
After getting the stroke renderer, we train the painter network on the training dataset (CelebA-HQ or ImageNet) for 2M iterations with batch size 32.}

(3) \ori{{\bf Train compositor network.} With the trained painter network and renderer, we train our compositor network with the DDPG framework on the training dataset (CelebA-HQ or ImageNet) for 2M iterations with batch size 32.}

\begin{table}[t]
\small
\centering
\setlength{\abovecaptionskip}{4pt}
\setlength{\belowcaptionskip}{-0.2cm}
\setlength\tabcolsep{3pt}
\renewcommand{\arraystretch}{1.2}
\caption{\hut{The quantitative comparison between the state-of-the-art methods and our model. 
% We conduct experiments on a 24G RTX3090 GPU, where Stylized Neural Painting \cite{zou2021stylized} can render at most 2,400 strokes due to excessive demand for memory, so there are no results for Stylized Neural Painting under 5,000 strokes.
}}
\scalebox{0.69}{
\begin{tabular}{cc|ccc|ccc}
\toprule
 Stroke&\multirow{2}{*}{Method}
 & \multicolumn{3}{c|}{ImageNet}& \multicolumn{3}{c}{CelebA-HQ}\\
 Num&&$\mathcal{L}_2$ Dist $\downarrow$&PSNR $\uparrow$&$\mathcal{L}_{PIPS}$ $\downarrow$&$\mathcal{L}_2$ Dist $\downarrow$&PSNR $\uparrow$&$\mathcal{L}_{PIPS}$ $\downarrow$\\
% Num                  &                            & $\mathcal{L}_2$ Dist $\downarrow$ & PSNR $\uparrow$ & $\mathcal{L}_{PIPS}$ $\downarrow$ & $\mathcal{L}_2$ Dist $\downarrow$ & PSNR $\uparrow$ & $\mathcal{L}_{PIPS}$ $\downarrow$ \\
\midrule
\multirow{7}{*}{200} & Paint Transformer \cite{liu2021paint}          & 0.0585          & 12.86          & 0.1984          & 0.0380          & 14.63          & 0.1738          \\
& Learning To Paint ~\cite{huang2019learning}                & \hut{0.0125}          & \hut{19.73}          & \hut{0.1636}          & \hut{0.0065}          & \hut{22.08}          & \hut{0.1578}          \\
& Semantic Guidance+RL \cite{singh2021combining}            & \hut{0.0191}          & \hut{17.67}          & \hut{0.1966}          & 0.0092          & 20.75          & \textbf{0.1176} \\
& Stylized Neural Painting \cite{zou2021stylized}           & \hut{0.0105}          & \hut{\textbf{20.73}}          & \hut{0.1625}          & \hut{0.0061}          & \hut{22.49}          & \hut{0.1584}          \\
& Parameterized Brushstrokes \cite{kotovenko2021rethinking} & \hut{0.0831}          & \hut{11.15}           & \hut{0.2259}          & \hut{0.0769}          & \hut{11.43}           & \hut{0.1927}          \\
& Im2oil~\cite{tong2022im2oil}&\hut{0.0331}&\hut{15.59}&\hut{0.1787}&\hut{0.0171}&\hut{18.15}&\hut{0.1822}\\
& Ours                       & \textbf{\hutnew{0.0102}} & \hutnew{20.42} & \textbf{\hutnew{0.1586}} & \textbf{0.0044} & \textbf{24.01} & 0.1324\\        
\midrule

\multirow{7}{*}{500} & Paint Transformer \cite{liu2021paint}          & 0.0379          & 14.80          & 0.1803          & 0.0227          & 16.89          & 0.1590          \\
& Learning To Paint \cite{huang2019learning}                 & \hut{0.0092}          & \hut{21.13}          & \hut{0.1453}          & \hut{0.0044}          & \hut{23.86}          & \hut{0.1449}          \\

& Semantic Guidance+RL \cite{singh2021combining}             &  \hut{0.0180}          & \hut{17.95}          & \hut{0.1970}          & 0.0087          & 21.00          & \textbf{0.1144} \\
& Stylized Neural Painting \cite{zou2021stylized}            & \hut{0.0088}          & \hut{21.26}          & \hut{0.1526}          & \hut{0.0046}         & \hut{23.65}          & \hut{0.1477}          \\
& Parameterized Brushstrokes \cite{kotovenko2021rethinking}  & \hut{0.0555}          & \hut{13.09}           & \hut{0.1681}          & \hut{0.0725}          & \hut{11.71}           & \hut{0.1897}          \\
& Im2oil~\cite{tong2022im2oil}&\hut{0.0263}&\hut{16.89}&\hut{0.1610}&\hut{0.0115}&\hut{19.83}&\hut{0.1409}\\
 & Ours                       & \textbf{\hutnew{0.0087}} & \textbf{\hutnew{21.52}} & \textbf{\hutnew{0.1430}} & \textbf{\hut{0.0035}} & \textbf{\hut{25.26}} & \hut{0.1202}     \\ 
\midrule
%1000 tensor(0.0035, device='cuda:0') tensor(25.2561, device='cuda:0') tensor(0.1202, device='cuda:0')

\multirow{7}{*}{1,000} & Paint Transformer \cite{liu2021paint}          & 0.0221          & 17.28          & 0.1622          & 0.0105          & 20.14          & 0.1443          \\
& Learning To Paint \cite{huang2019learning}                 & \hut{0.0076}          & \hut{21.99}          & \hut{0.1385}          & \hut{0.0034}          & \hut{24.94}          & \hut{0.1327}      \\

& Semantic Guidance+RL \cite{singh2021combining}            & \hut{0.0171}          & \hut{18.16}          & \hut{0.1953}          & 0.0069          & 22.10          & 0.1201          \\
& Stylized Neural Painting \cite{zou2021stylized}           & \hut{0.0079}          & \hut{21.90}          & \hut{0.1458}          & \hut{0.0045}          & \hut{23.73}          & \hut{0.1411}          \\
& Parameterized Brushstrokes \cite{kotovenko2021rethinking} & \hut{0.0502}          & \hut{13.56}           & \hut{0.1524}          & \hut{0.0692}          & \hut{11.94}           & \hut{0.1884}          \\
& Im2oil~\cite{tong2022im2oil}&\hut{0.0195}&\hut{18.11}&\hut{0.1452}&\hut{0.0058}&\hut{22.99}&\hut{0.1506}\\
& Ours                       & \textbf{\hutnew{0.0068}} & \textbf{\hutnew{22.72}} & \textbf{\hutnew{0.1305}} & \textbf{\hut{0.0024}} & \textbf{\hut{26.62}} & \textbf{\hut{0.1060}} \\
\midrule
%1000 tensor(0.0024, device='cuda:0') tensor(26.6197, device='cuda:0') tensor(0.1060, device='cuda:0')

\multirow{7}{*}{\hut{3,000}} & Paint Transformer \cite{liu2021paint}          &\hut{0.0135}          & \hut{19.39}          & \hut{0.1375}          & \hut{0.0067}         & \hut{22.16}          & \hut{0.1709}          \\
& Learning To Paint \cite{huang2019learning}                & \hut{0.0064}          & \hut{22.71}          & \hut{0.1278}          & \hut{0.0029} & \hut{25.72}          & \hut{0.1219}          \\

& Semantic Guidance+RL \cite{singh2021combining}            & \hut{0.0160}          & \hut{18.34}          & \hut{0.1965}          & 0.0073          & 21.78          & 0.1201          \\
& Stylized Neural Painting \cite{zou2021stylized}           & \hut{0.0100}          & \hut{20.62}          & \hut{0.1441}          & \hut{0.0070}          & \hut{21.73}          & \hut{0.1434}          \\
& Parameterized Brushstrokes \cite{kotovenko2021rethinking} & \hut{0.0437}          & \hut{14.18}           & \hut{0.1385}          & \hut{0.0617}          & \hut{12.50}           & \hut{0.1850}          \\
& Im2oil~\cite{tong2022im2oil}&\hut{0.0119}&\hut{20.48}&\hut{0.1220}&\hut{0.0029}&\hut{26.07}&\hut{0.1295}\\
& Ours                       & \textbf{\hutnew{0.0052}} & \textbf{\hutnew{23.95}} & \textbf{\hutnew{0.1106}} & \textbf{0.0016} & \textbf{28.33} & \textbf{0.0839} \\
\midrule

\multirow{7}{*}{5,000} & Paint Transformer \cite{liu2021paint}          &  \hut{0.0128}         & \hut{19.64}          & \hut{0.1353}          & \hut{0.0062}          & \hut{22.47}          & \hut{0.1692}          \\
& Learning To Paint \cite{huang2019learning}               & \hut{0.0061}          & \hut{22.90}          & \hut{0.1255}          & \hut{0.0028}          & \hut{25.88}          & \hut{0.1193}          \\

& Semantic Guidance+RL \cite{singh2021combining}           & \hut{0.0161}          & \hut{18.33}          & \hut{0.1950}          & 0.0075          & 21.84          & 0.1196          \\
& Stylized Neural Painting \cite{zou2021stylized}          & \hut{0.0081}          & \hut{21.29}         & \hut{0.1379}               & \hut{0.0055}               & \hut{22.63}              & \hut{0.1667}               \\
& Parameterized Brushstrokes \cite{kotovenko2021rethinking} & \hut{0.0400}          & \hut{14.58}           & \hut{0.1332}          & \hut{0.0585}          & \hut{12.82}           & \hut{0.1847}          \\
& Im2oil~\cite{tong2022im2oil}&\hut{0.0091}&\hut{21.61}&\hut{0.1118}&\hut{0.0021}&\hut{27.23}&\hut{0.1195}\\
& Ours                       & \textbf{\hutnew{0.0046}} & \textbf{\hutnew{24.57}} & \textbf{\hutnew{0.1026}} & \textbf{0.0014} & \textbf{28.79} & \textbf{0.0820} \\ 
\bottomrule
\end{tabular}}
\label{tab:quantitative-comparison}
\vspace{-0.2in}
\end{table}

\subsection{Image To Painting (Reconstruction)}
\begin{figure}[t]
\centering
\includegraphics[width=0.46\textwidth]{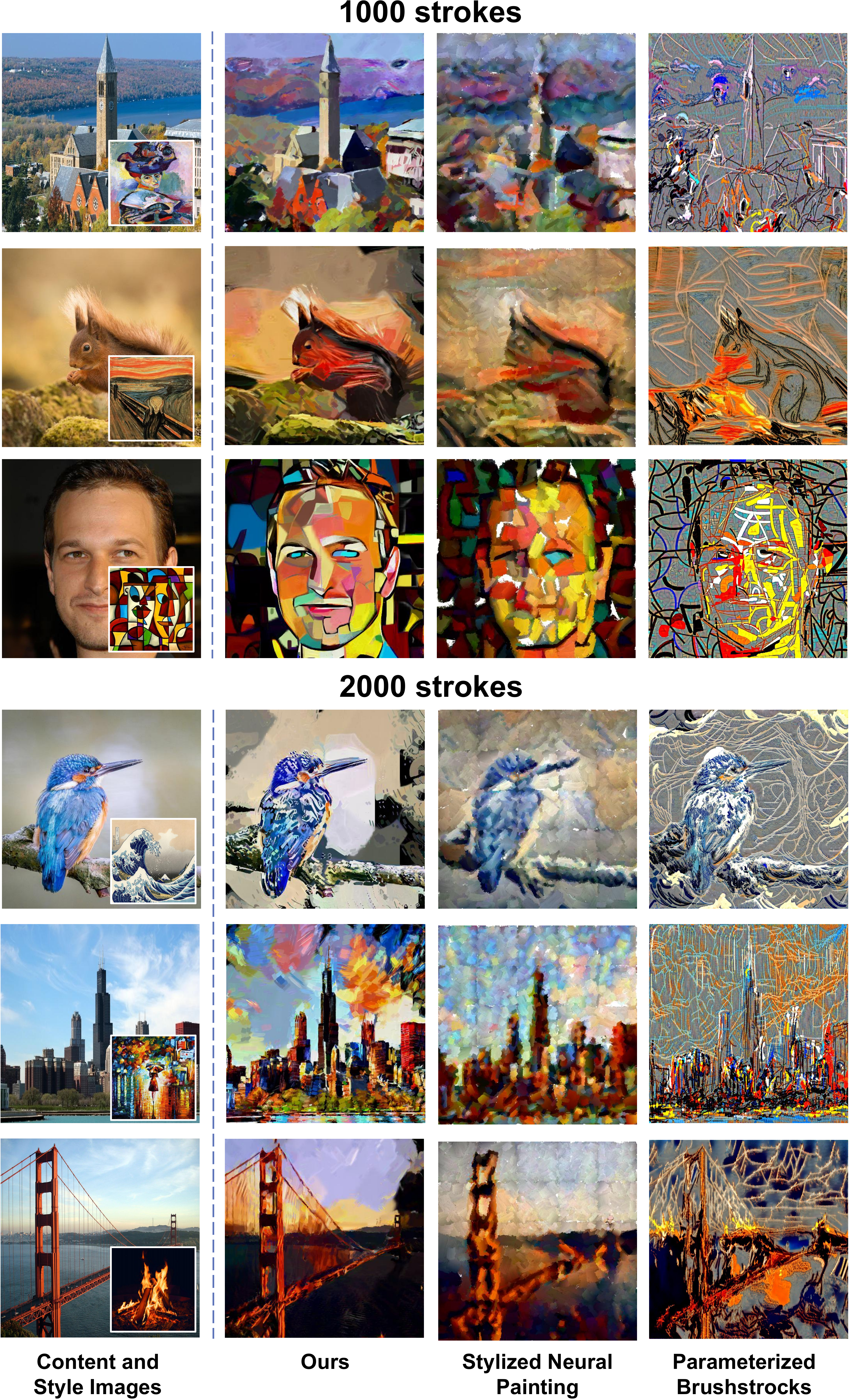}
\vspace{-0.1in}
\caption{\hut{Stroke-based style transfer comparison with Stylized Neural Painting ~\cite{zou2021stylized} and Parameterized Brushstrokes ~\cite{kotovenko2021rethinking} under 1,000 and 2,000 strokes.}}
\label{fig:comparison on SBR stylization}
\vspace{-0.15in}
\end{figure}

{\bf \hutnew{Experiment Setting.}} The State-of-the-art model can be classified into the RL-based model (Learning To Paint ~\cite{huang2019learning} and Semantic Guidance+RL ~\cite{singh2021combining}), the \hut{DL-based} model (Paint Transformer ~\cite{liu2021paint}), the optimization-based model (Stylized Neural Painting ~\cite{zou2021stylized} and Parameterized Brushstrokes ~\cite{kotovenko2021rethinking}) \hut{and the traditional search-based model (Im2Oil~\cite{tong2022im2oil})}. 
\hutnew{In this section, we compare our model to these methods~\cite{huang2019learning, singh2021combining, liu2021paint,kotovenko2021rethinking,zou2021stylized,tong2022im2oil}. 
% we implement both the uniform-block-dividing strategy and the boundary inconsistency reduction strategies utilized by the state-of-the-art models, which include the use of 
For a fair comparison, we use \textbf{the same oil painting brushstrokes } for all the methods  except Parameterized Brushstrokes since its key contribution is the specially designed renderer for Bézier stroke and follow the official implementation of all the methods. We compare \hut{the rendered results} with 200, 500, 1,000, \hut{3,000} and 5,000 strokes respectively and evaluate the results both qualitatively and quantitatively.}

% Among them, 
% Learning To Paint ~\cite{huang2019learning} is the first model which can reconstruct most of the details of an input image. Based on it, Semantic Guidance+RL~\cite{singh2021combining} introduce\hut{s} the semantic guidance into the RL training process which can focus more on the semantic part. To avoid the long training time and the unstable agent of the RL method, Paint Transformer~\cite{liu2021paint} propose a RL free model for SBR task. However, the painter network in Paint Transformer can only reconstruct the image in a coarse view and need much more strokes than the RL-based models to reconstruct the same image. Besides, Stylized Neural Painting and Parameterized Brushstrokes render the images based on the optimization process which suffers from a long optimization time for each single image. \hut{Among all the methods, Im2Oil~\cite{tong2022im2oil} is the only method that employs traditional greedy-search strategy, which also suffers from the long searching time.}
\hut{Among the existing methods, Learning To Paint~\cite{huang2019learning} is the pioneering model capable of accurately reconstructing most of the details in the input image. Building on it, Semantic Guidance+RL~\cite{singh2021combining} incorporates semantic guidance into the RL training process, enabling the model to focus more on the semantic part. To overcome the challenges posed by the long training time and unstable agent of the RL method, Paint Transformer~\cite{liu2021paint} proposes a RL-free model. However, the painter network in Paint Transformer only reconstructs images at a coarse level and requires much more strokes than the RL-based models for the same image. Additionally, both Stylized Neural Painting and Parameterized Brushstrokes render images through the optimization process, which suffers from a long optimization time for each individual image. Among all these methods, Im2Oil~\cite{tong2022im2oil} is the only approach that adopts the traditional greedy-search strategy, which also entails considerable search time.}
% In this section, we compare our model with the state-of-the-art models~\cite{huang2019learning,singh2021combining,liu2021paint} in real image rendering  \hut{where
% we implement the state-of-the-art models with both the uniform-block-dividing strategy and their strategies in decreasing the boundary inconsistency problem (i.e., Paint Transformer use additional strokes at the end of the painting process, Learning To Paint averages the pixels at the boundary and Stylized Neural Painting employs progressive painting strategy.)}. 

Figure \ref{fig:image to painting} shows the rendered images by Learning To Paint, Semantic Guidance+RL, Paint Transformer, Stylized Neural Painting and Parameterized Brushstrokes \hut{and Im2Oil}. It can be seen that all the results generated by the uniform-block-dividing strategy (Learning To Paint, Paint Transformer and Stylized Neural Painting) have the boundary inconsistency problem. Moreover, the Semantic Guidance+RL method cannot use the uniform-block-dividing strategy so that it fails to reconstruct the details in a complex real image. \hut{Im2Oil has a relatively good visual quality, but lacks some details especially when the stroke number is limited (e.g., 200 strokes).} In contrast, our model not only solves the boundary inconsistency artifacts, but also render the images with the richest details. We also conduct a quantitative comparison between our model and the state-of-the-art methods. We evaluate the average $L_2$ distance, PSNR and $\mathcal{L}_{PIPS}$ between 1,000 target images and the generated images among these methods in Table \ref{tab:quantitative-comparison} (\hut{100} images for Stylized Neural Painting, Parameterized Brushstrokes \hut{and Im2Oil} due to their slow optimization process) . It shows that our model outperforms all the state-of-the-art methods in most of the conditions.

\subsection{Stroke-based Style Transfer}
% We first compare the effectiveness between our DT loss and the existing DT loss~\cite{pham2020differentiable}. We randomly sampled 5000 images $I$ with their edge map$E_t$ in the dataset and use our painter to render these images into an edge map $E_r$ too. We compare the accuracy of the two DT loss with the the true nondifferentiable DT loss begin the standard answer. Table \ref{tab:comparison on the DT loss} shows the compared results. We can see that our DT loss outperforms the existing DT loss in the SBR task.

We compare our model with the state-of-the-art stroke-based style transfer methods \hut{Stylized Neural Painting~\cite{zou2021stylized} and Parameterized Brushstrokes~\cite{kotovenko2021rethinking} using the same number of strokes (1,000 and 2,000) and \hutnew{their} default stylization settings.  \hutnew{The results in Fig.~\ref{fig:comparison on SBR stylization}} demonstrate that Stylized Neural Painting suffers from severe boundary artifacts, while Parameterized Brushstrokes fails to maintain most of the contents. Furthermore, both methods lose a significant amount of information in the object boundaries, resulting in blurred edges and structures in the generated images. (Note that Parameterized Brushstrokes~\cite{kotovenko2021rethinking} can achieve relatively good stylization results with a large amount stroke count (over 10,000) as shown in their paper) In contrast, our model not only preserves the structure of content images well, but also delivers superior stylization effects and visual quality when compared to the aforementioned methods. Furthermore, we quantitatively compare the stylization effects of each method by computing the distance between the gram matrices  of the generated images and style images in Fig.~\ref{fig:comparison on SBR stylization} as a measurement of stylization effectiveness. The results explicitly show that our method has a significantly improved average distance of \textbf{0.6943} compared to Stylized Neural Painting (1.4048) and Parameterized Brushstrokes (4.0274), indicating that our model has a superior stylization performance. }

\subsection{Ablation Study}
\label{ssec:ablation}

\begin{figure}[t]
\centering
\includegraphics[width=0.45\textwidth]{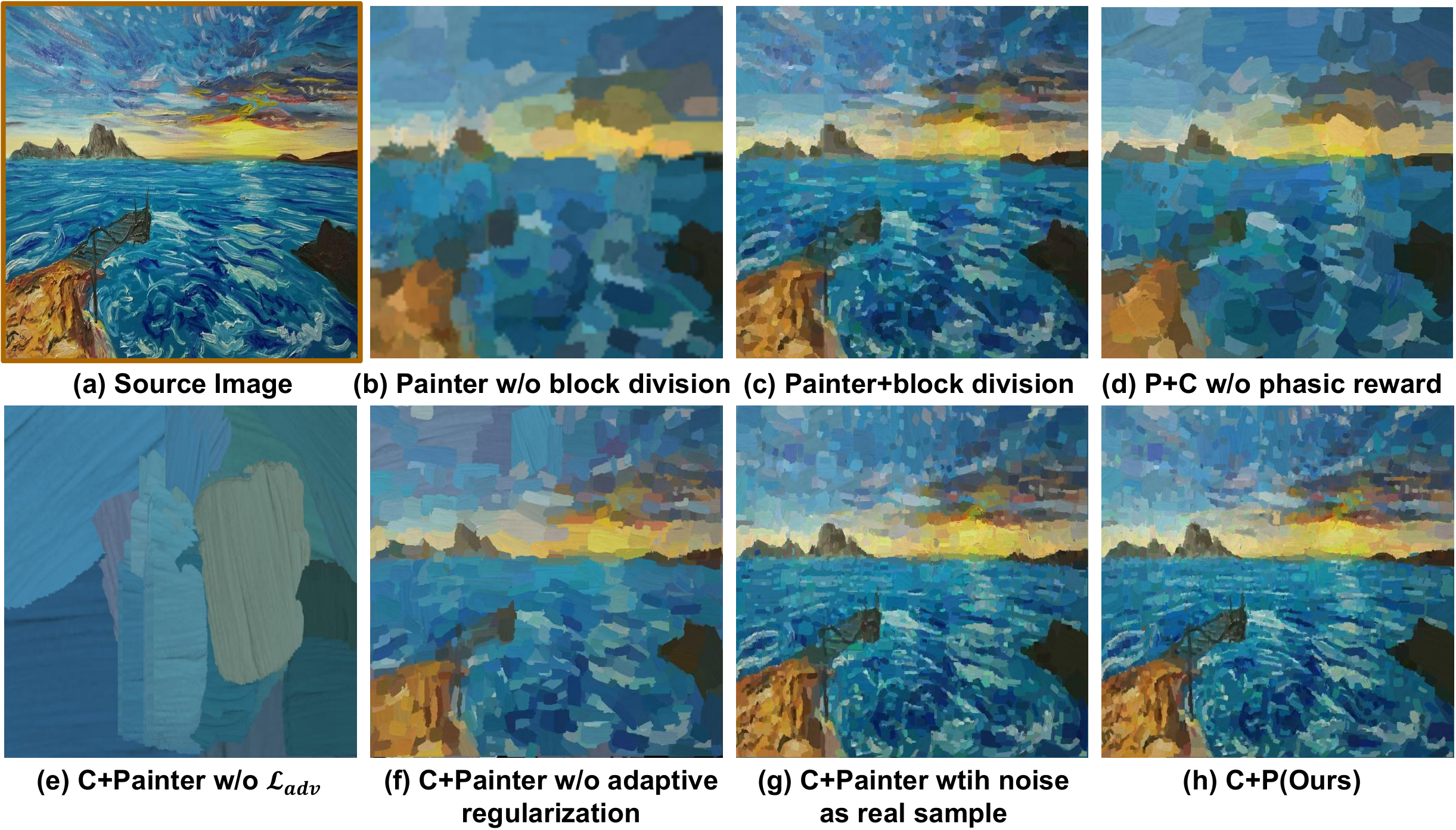}
\vspace{-0.1in}
\caption{\hutnew{Ablation Study on the compositor (C) and painter network (P). The model without compositor either fails to reconstruct the details (b) or suffers from the boundary artifacts (c). The model without phasic reward reconstructs less details (d) than the standard model (h). As for the painter network, the model without $\mathcal{L}_{adv}$ (e) tends paint the same big strokes repeatedly. And the model without the adaptive regularization factor $\gamma$ (f) fails to paint the image details. The model taking random noise as the \hutnew{real} WGAN sample (g) has a similar results to the standard model (h). }}
\label{fig:ablation study on PC}
\vspace{-0.1in}
\end{figure}

{\bf Ablation study on the compositor network.} In this part, we validate the importance of our compositor network and the  function of our phasic reward mechanism. We train 3 ablated models: (1) a model without the compositor network (only a painter network and a stroke renderer); (2) a model without the compositor network but using the uniform-block-dividing strategy; (3) an ablated model without the phasic reward. 
Tab. \ref{tab:Ablation Study} and Fig.~\ref{fig:ablation study on PC} show the comparison results. \hut{It can be seen that the model without the compositor network or phasic reward fails to reconstruct image details and the model with uniform-block-dividing strategy suffers from the boundary artifacts (Fig.~\ref{fig:ablation study on PC}(c)).}

\begin{table}[t]
\small
\centering
\setlength{\abovecaptionskip}{4pt}
\setlength{\belowcaptionskip}{-0.2cm}
\setlength\tabcolsep{3pt}
\renewcommand{\arraystretch}{1.2}
\caption{\hutnew{Ablation Study on the compositor and painter network on the ImageNet dataset.}}
\scalebox{0.65}{
\begin{tabular}{cc|ccc|ccc}
\toprule
\multicolumn{2}{c|}{Method}
& \multicolumn{3}{c|}{\hutnew{1,000} Strokes}& \multicolumn{3}{c}{5,000 Strokes}\\
Compositor&Painter&$\mathcal{L}_2$ Dist $\downarrow$&PSNR $\uparrow$&LPIPS $\downarrow$ &$\mathcal{L}_2$ Dist $\downarrow$&PSNR $\uparrow$&LPIPS $\downarrow$\\
\midrule
no compositor&our painter&\hutnew{0.0122}&\hutnew{19.88}&\hutnew{0.1739}&\hutnew{0.0134}&\hutnew{19.45}&\hutnew{0.1705}\\

w/o phasic reward&our painter&\hutnew{0.0099}&\hutnew{20.31}&\hutnew{0.1389}&0.0098&20.83&0.1364\\
block division&our painter&\hutnew{0.0071}&\hutnew{22.32}&\hutnew{0.1345}&0.0058&23.12&0.1194\\
\midrule
our compositor& w/o $\mathcal{L}_{adv}$&\hutnew{0.0241}&\hutnew{16.62}&\hutnew{0.1766}&0.0239&16.65&0.1762\\
our compositor& w/o adaptive regularization &\hutnew{0.0129}&\hutnew{19.47}&\hutnew{0.1689}&\hutnew{0.0109}&\hutnew{20.20}&\hutnew{0.1558}\\
our compositor& noise as \hutnew{real} sample &\hutnew{0.0072}&\hutnew{22.35}&\hutnew{0.1325}&0.0049&24.02&0.1047\\
\midrule
our compositor&our painter&\textbf{\hutnew{0.0068}}&\textbf{\hutnew{22.72}}&\textbf{\hutnew{0.1305}}&\textbf{0.0046}&\textbf{24.54}&\textbf{0.1026}\\
\bottomrule
\end{tabular}}
\label{tab:Ablation Study}
%\vspace{-0.2in}
\end{table}

{\bf Ablation study on the painter network.}
To validate the effectiveness of the key components of our painter network, %: the adversarial loss $\mathcal{L}_{adv}$ and the adaptive regularization factor $\gamma$. 
we train 3 ablated models: one model without the adversarial loss $\mathcal{L}_{adv}$, one model without the adaptive regularization factor $\gamma$, and one model with random noises being the \hutnew{real} samples. \hut{We employ both the compositor and painter network to conduct the experiments and the results are shown in} Tab. \ref{tab:Ablation Study} and Fig.~\ref{fig:ablation study on PC}. It can be seen that the model without $\mathcal{L}_{adv}$ tends to  \hut{ paint the same big strokes repeatedly. And the model without $\gamma$ suffers from an unstable training process, resulting in a worse painting performance than our model.} Moreover, with random noises being the \hutnew{real} samples, the model can also achieve a similar performance as our baseline model, indicating that %the key component that making our model feasible is the penalizing function of the seen \hutnew{fake} samples by the discriminator $D$.
penalizing seen \hutnew{fake} samples to avoid painting similar strokes is the main function of our discriminator $D$.

% {\bf Ablation study on DT loss.} \hut{In this section, we perform an ablation study on the DT loss of our stroke-based stylization method in order to assess its efficacy. As demonstrated in Fig.~\ref{fig:ablation study on dt loss}, the model lacking DT loss exhibits reduced capability in preserving the structural consistency of the content image. In contrast, our model incorporating DT loss maintains the structure of the content image effectively.}

% \begin{figure}[t]
% \centering
% \includegraphics[width=0.45\textwidth]{figures/ablation study on dt loss.pdf}
% \vspace{-0.1in}
% \caption{\hut{Ablation study on distance transform loss in stroke-based stylization. The model without DT loss loses some structure information (shown in the green boxes). In contrast, our model with the DT loss preserves the structures well.}}
% \label{fig:ablation study on dt loss}
% \vspace{-0.15in}
% \end{figure}

\section{Conclusion}
In this paper, we propose {\it Compositional Neural Painter}, a novel stroke-based rendering framework which dynamically predicts the next painting region based on the current canvas, instead of dividing the image plane uniformly into $K\times K$ painting regions. 
% Our painting process can be decomposed into several steps, where at each step, a compositor network trained with a phasic RL strategy first predicts the next painting region, then a painter network trained with a WGAN discriminator predicts stroke parameters, and a stroke renderer paints the strokes onto the painting region of the current canvas. 
Moreover, we extend our method to \hutnew{edge-promoting} stroke-based style transfer \hutnew{task} with a novel differentiable distance transform loss. 
Extensive \hut{quantitative and qualitative} experiments show that our model outperforms the state-of-the-art methods in both stroke-based neural rendering and stroke-based stylization.


\begin{acks}
 This work was supported by National Natural Science Foundation of China (72192821, 62272447, 61972157), Shanghai Sailing Program (22YF1420300), Shanghai Municipal Science and Technology Major Project (2021SHZDZX0102), Shanghai Science and Technology Commision (21511101200), CCF-Tencent Open Research Fund (RAGR20220121), Young Elite Scientists Sponsorship Program by CAST (2022QNRC001), Beijing Natural Science Foundation (L222117), the Fundamental Research Funds for the Central Universities (YG2023QNB17).
\end{acks}

\bibliographystyle{ACM-Reference-Format}
\balance
\bibliography{sample-base}


\clearpage

\appendix

\section{Overview}
\par This appendix consists of:

1) The training details of our model (Sec.~\ref{sec:training details});

2) Ablation study on our distance transform loss (Sec.~\ref{sec:ablation study});

3) Visualization of the painting process of our model (Sec.~\ref{sec:painting process});

4) User study on the painting performance (Sec.~\ref{sec:user study});

5) Comparison between the painter networks of our model and the existing methods (Sec.~\ref{Sec:Comparison between the painter networks});

6) Comparison with the local attention window proposed in Intelli-Paint~\cite{singh2022intelli} (Sec.~\ref{sec:comparison with intelli-paint});

7) More comparison results with the existing methods on stroke-based painting (Sec. \ref{Sec:More Comparison on Stroke-based Painting});

8) More comparison results on Stylization which includes the stylization method in pixel level (Sec. \ref{Sec:More experiments on Stroke-based Stylization});

9) The time complexity analysis on each method (Sec.~\ref{Sec:time complexity}). 

\section{Training Details}
\label{sec:training details}
\ori{The training process consists of three steps:}

(1) \ori{{\bf Train stroke renderer.} We first train our stroke renderer with synthesized data. In detail, we randomly sample stroke parameters $s$ and transform the basic stroke to get the target stroke mask \hut{$\hat{M}_s$}. With the data pair \hut{$(s,\hat{M}_s)$}, we train our renderer by minimizing Eq.(7) in the main paper for 1M iterations with batch size 32.}

(2) \ori{{\bf Train painter network.}
After getting the stroke renderer, we train the painter network on the training dataset (CelebA-HQ or ImageNet) for 2M iterations with batch size 32.}

(3) \ori{{\bf Train compositor network.} With the trained painter network and renderer, we train our compositor network with the DDPG framework on the training dataset (CelebA-HQ or ImageNet) for 2M iterations with batch size 32.}

\section{Ablation study on DT loss}
\label{sec:ablation study}
\begin{figure}[t]
\centering
\includegraphics[width=0.45\textwidth]{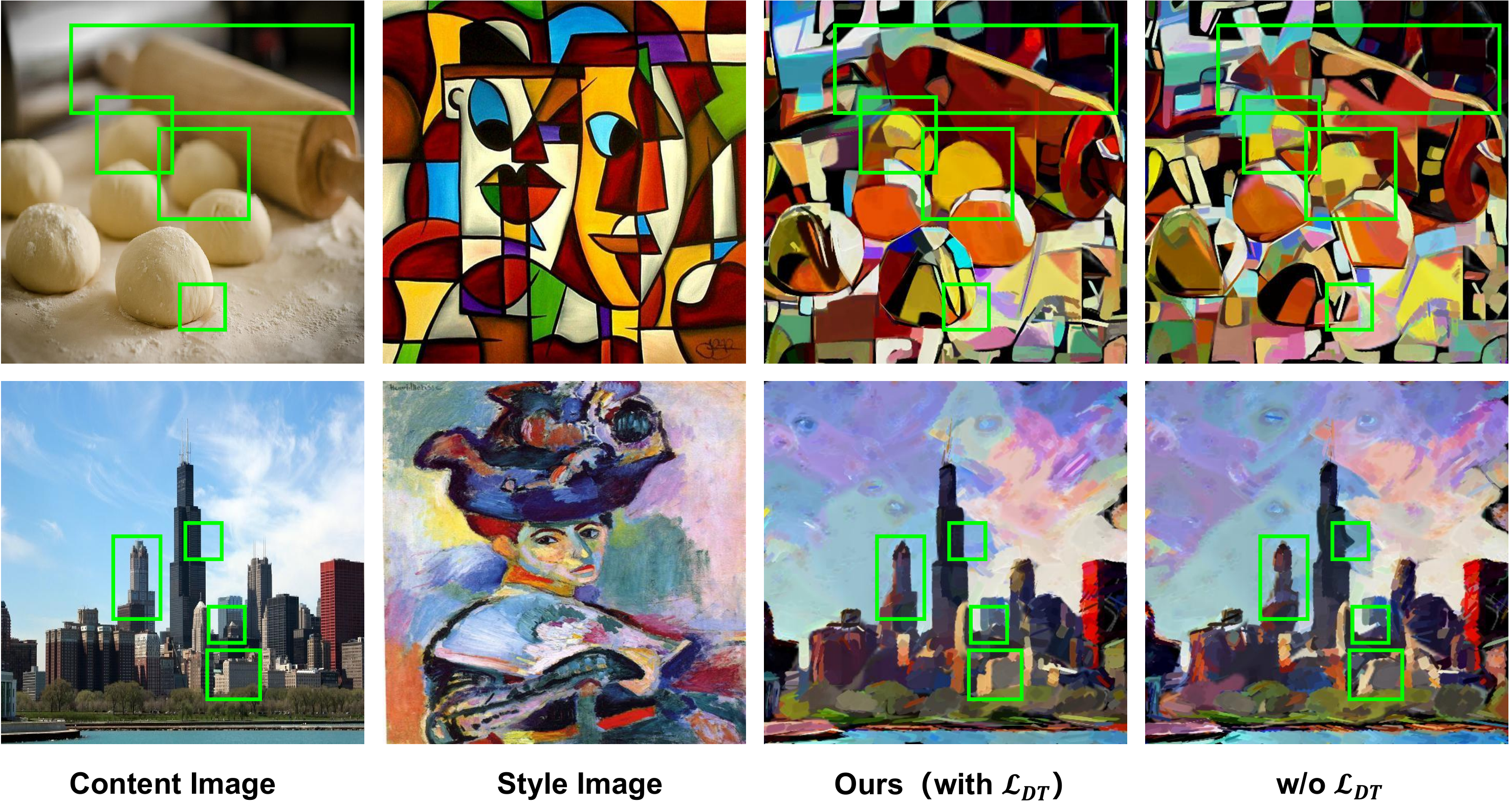}
%\vspace{-0.1in}
\caption{\hut{Ablation study on distance transform loss in stroke-based stylization. The model without DT loss loses some structure information (shown in the green boxes). In contrast, our model with the DT loss preserves the structures well.}}
\label{fig:ablation study on dt loss}
%\vspace{-0.15in}
\end{figure}

\hut{In this section, we perform an ablation study on the distance transform (DT) loss of our stroke-based stylization method in order to assess its efficacy. We conducted a comparative experiment between our stroke-based stylization model (with DT loss) and the DT loss-free model as demonstrated in Fig.~\ref{fig:ablation study on dt loss}. It can be observed that in the first row of the figure, the DT loss-free model loses a significant amount of edge information in regions with comparatively blurred edges. Moreover, in the second row of the figure, the DT loss-free model exhibits edge disturbances specifically along straight edges (i.e., building outlines). In contrast, our model incorporating DT loss maintains the structure of the content image effectively in both the situations (blurred edges and straight edges).}

\section{Painitng Process}
\label{sec:painting process}

\begin{figure*}[t]
\centering
\includegraphics[width=0.9\textwidth]{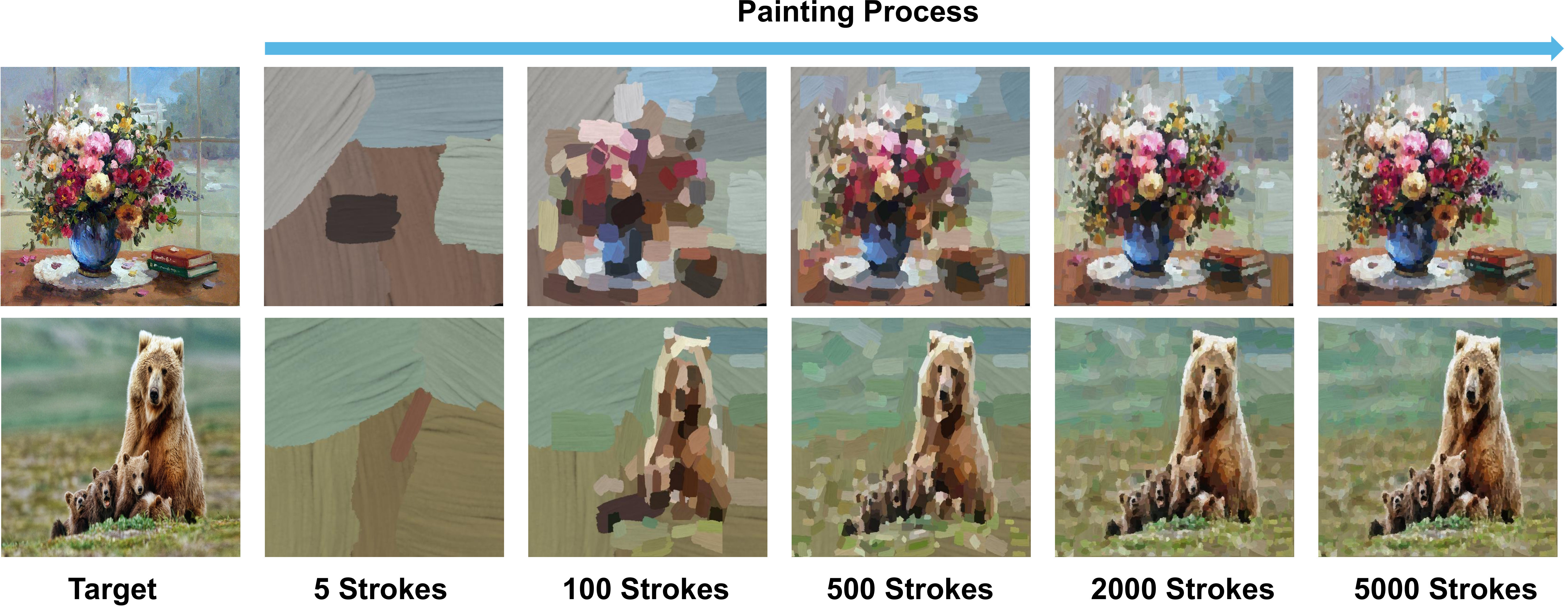}
%\vspace{-0.1in}
\caption{The Painting process of our model. Our model initiates the painting process by capturing the overall structure of the entire image and subsequently enriches the artwork with finer details. As the number of brush strokes increases, the painting results become increasingly intricate and refined.
}
\label{fig:painting process}
%\vspace{-0.1in}
\end{figure*}

In this section, we visualize the painting process of our model, as shown in Fig.~\ref{fig:painting process}. We present the images generated by the model at various stages of painting, corresponding to 5, 100, 500, 2000, and 5000 brush strokes, respectively. It can be observed that our model initially tends to capture the overall background contours of the image during the early stages of painting. Subsequently, it gradually enriches the image with finer details. As the number of brush strokes increases, the model's capability to reconstruct intricate details strengthens, ultimately yielding satisfying results in the final painting.

\section{User study}
\label{sec:user study}
To further validate the effectiveness of our model, we invite 30 volunteers to rank the generated results of our approach, Learning To Paint~\cite{huang2019learning}, Paint Transformer~\cite{liu2021paint}, Stylized Neural Painting~\cite{zou2021stylized} and Im2Oil~\cite{tong2022im2oil}, based on their perceived quality from best to worst. The results are presented in Tab.~\ref{tab:user study}. It can be seen that our model gets the top rank in 65\% of the cases. Furthermore, our average ranking stands at an impressive 1.56, indicating superior performance compared to the other methods.

% {\bf comparison with pixel loss.}
% We first compare our DT loss with pixel loss in Figure \ref{fig:pixel loss vs DT loss} by optimizing the strokes' parameters.  we aims to cover the edge in target image with the stroke edges. When starting from a source stroke edge map which do not overlap with the target edge, pixel loss fails to cover the target edge while our DT loss can cover it well.

% \begin{figure}[t]
% \centering
% \includegraphics[width=0.45\textwidth]{figures/pixel loss vs DT loss.pdf}
% \caption{The comparison between pixel loss and DT loss in edge matching by parameter optimization. The pixel loss is not differentiable when the edges are not overlapped while our DT loss can be get a good performance.}
% \label{fig:pixel loss vs DT loss}
% %\vspace{-0.1in}
% \end{figure}

\section{Comparison between the painter networks}
\label{Sec:Comparison between the painter networks}

The existing painter networks can be divided into two types: one that utilizes reinforcement learning (Learning To Paint ~\cite{huang2019learning}, Semantic Guidance+RL~\cite{singh2021combining}, etc), and the other that does not (Paint Transformer~\cite{liu2021paint}). 
(a) Compared to Painter Transformer that does not employ reinforcement learning, our approach trains the network using real data rather than synthetic data, and incorporates adversarial learning to enhance the exploration capability, resulting in significantly improved painting performance. 
(b) Compared to reinforcement learning-based methods, our approach adopts deep learning training strategy, which has a simpler training framework and yields a stabler training process, while achieving better stroke-based painting results.

To validate this, we compare the painter networks of Learning To Paint, Paint Transformer, Semantic Guidance+RL and ours without block division on ImageNet~\cite{deng2009Imagenet} and CelebA-HQ datasets~\cite{karras2018progressive} (200 strokes) in Tab. ~\ref{tab:quantitative comparison on painter}. It can be seen that our painter network exhibits the lowest $\mathcal{L}_2$ distance and LPIPS score, as well as the highest PSNR, indicating its superior image reconstruction capability over all other models.

\section{Comparison with Intelli-Paint}
\label{sec:comparison with intelli-paint}
\hut{In Intelli-Paint~\cite{singh2022intelli}, the concept of attention window is proposed. Initially, Intelli-Paint employs object detection method on the input image to derive a global attention window. Using this global attention window, a neural network is trained to simultaneously predict both the local attention window and stroke parameters. Subsequently, the stroke parameters are mapped onto the local attention window for painting, which is reminiscent of our dynamic painting region strategy. Nevertheless, the motivation, usage of attention window, and resulting painting effects between Intelli-Paint and our model are significantly different.}

\hut{From a motivation perspective, Intelli-Paint uses the attention window to simulate the painting process of human artists, where the $i$th stroke and $i+1$th stroke are typically drawn close to each other in the canvas. They constantly slide the attention window, which enables the model to paint anywhere in the detected object window. In contrast, our goal is to utilize the painting region (can be regarded as a type of attention window) to enable the painter network to focus solely on the local details of the image, which leads to better reconstruction 
of image details. }

\begin{table}[t]
\caption{User study results summarized from rankings of 30 participants. Our model ranks first in 65\% cases with an average ranking of 1.56.}
\footnotesize
\centering
\setlength{\abovecaptionskip}{4pt}
\setlength{\belowcaptionskip}{-0.2cm}
\setlength\tabcolsep{4pt}
\renewcommand{\arraystretch}{1.2}
\scalebox{0.9}{
\begin{tabular}{cccccc}
\toprule
Method &Ours &\makecell[c]{Learning \\To Paint} &\makecell[c]{Paint \\ Transformer} &\makecell[c]{Stylized Neural\\ Painting}& Img2Oil\\
\midrule
\makecell[c]{Percentage of \\ ranking  first
(\%)$\uparrow$}&\textbf{65.00\%}&14.89\%&5.78\%& 9.56\%& 4.78\% \\
Average ranking$\downarrow$&\textbf{1.56}&2.54&3.55&3.18&4.15\\
\bottomrule
\end{tabular}}
\label{tab:user study}
\end{table}

\begin{table}[t]
\small
\centering
\setlength{\abovecaptionskip}{4pt}
\setlength{\belowcaptionskip}{-0.2cm}
\setlength\tabcolsep{3pt}
\renewcommand{\arraystretch}{1.2}
\caption{Comparison between the painter networks of Learning To Paint~\cite{huang2019learning}, Semantice Guidance+RL~\cite{singh2021combining}, Paint Transformer~\cite{liu2021paint} and ours on ImageNet~\cite{deng2009Imagenet} and Celeba-HQ~\cite{karras2018progressive} dataset with 200 strokes. It can be seen that our painter network outperforms the other ones in stroke-based rendering.}
\scalebox{0.8}{
\begin{tabular}{c|ccc|ccc}
\toprule
 \multirow{2}{*}{Method}
 & \multicolumn{3}{c|}{ImageNet}& \multicolumn{3}{c}{CelebA-HQ}\\
 &$\mathcal{L}_2$ Dist $\downarrow$&PSNR $\uparrow$&LPIPS $\downarrow$&$\mathcal{L}_2$ Dist $\downarrow$&PSNR $\uparrow$&LPIPS $\downarrow$\\
\midrule
Learning To Paint              & 0.0160          & 18.48          & 0.1871          & 0.0082          & 21.17          & 0.1388          \\
Semantic Guidance+RL           & 0.0197          & 17.65          & 0.2101          & 0.0092          & 20.75          & \textbf{0.1176} \\
Paint Transformer        & 0.1497          &   8.54        &   0.2180        &  0.1127         &    10.06      &    0.2045       \\
Our Painter                       & \textbf{0.0108} & \textbf{20.12} & \textbf{0.1639} & \textbf{0.0064} & \textbf{22.19} & 0.1299\\        
\bottomrule
\end{tabular}}
\label{tab:quantitative comparison on painter}
\end{table}

\hut{In terms of usage, Intelli-Paint employs a single network to simultaneously predict both the attention window and stroke parameters, which poses a significant burden on the RL-based neural painting model. Similar to Learning To Paint~\cite{huang2019learning}, the network in Intelli-Paint finds it challenging to predict a long sequence of stroke parameters, such as those exceeding 2,000 strokes. In contrast, our approach separates the prediction of the painting region and stroke parameters into two distinct stages, which are trained separately by combining the compositor network and the painter network. This two-stage training and testing strategy effectively enhances the number of strokes in the painting process.}

\begin{figure}[t]
\centering
\includegraphics[width=0.48\textwidth]{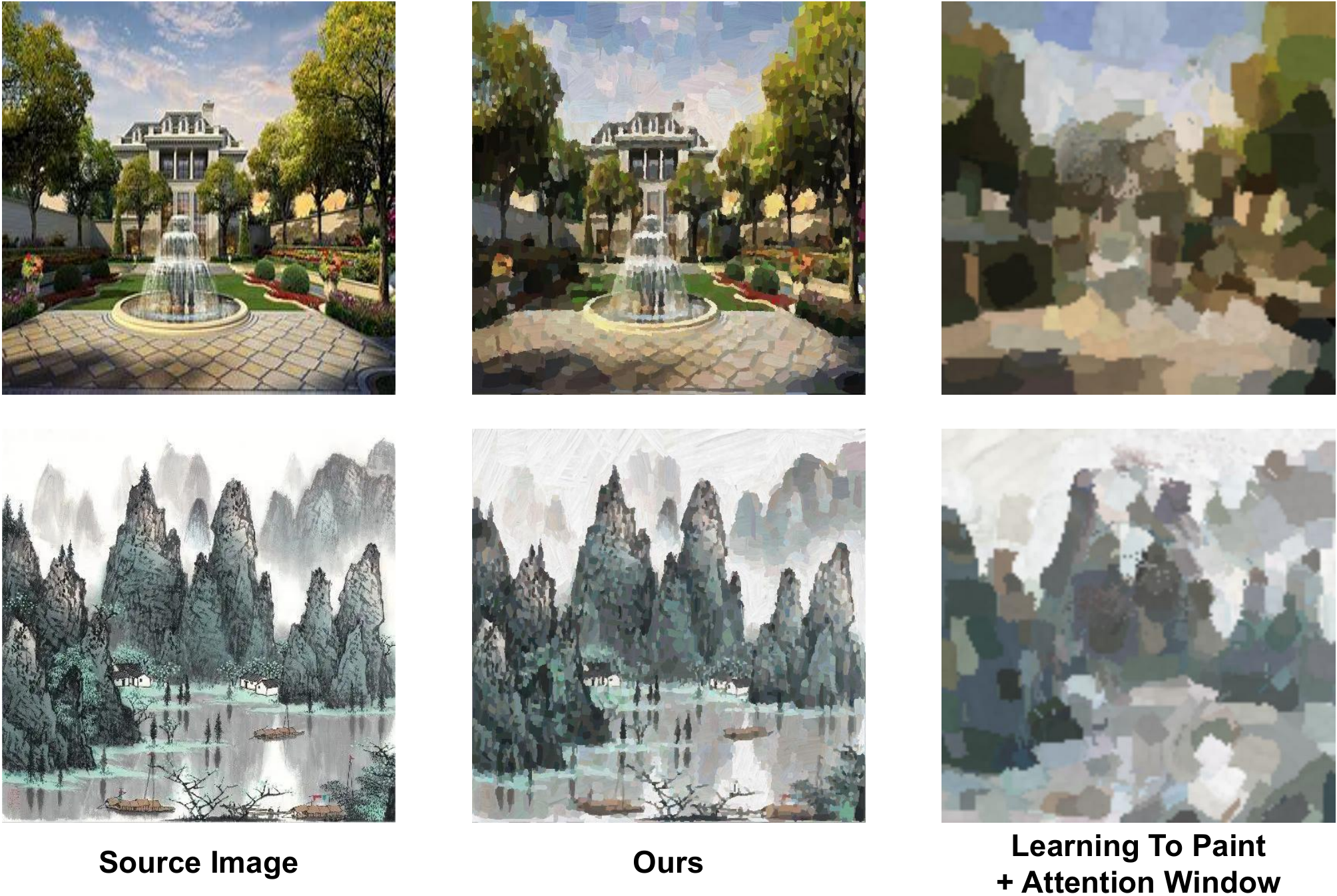}
\caption{The comparison between our methods and the reproduction of attention window in Intelli-Paint. It shows that the attention window in Intelli-Paint can not reconstruct the image details.
}
\label{fig:compare with Intelli-Paint}
%\vspace{-0.1in}
\end{figure}

\hut{Finally, in terms of resulting effects, Intelli-Paint focuses on painting images with clear objects and has difficulty reconstructing satisfactory details for images without clear objects, such as landscapes, due to limitations in its network design (one network predicts both the attention window and stroke parameters at the same time). In contrast, our model can paint images with fine-grained details in any situation. As Intelli-paint's code is not open-sourced, we replicate their attention window design based on Learning To Paint~\cite{huang2019learning} since both of them are trained with DDPG and demonstrate that their attention window design could not achieve accurate reconstruction of image details but only mimic the human painting style, namely, the close proximity between two consecutive brushstrokes. The comparison results are shown in Fig.~\ref{fig:compare with Intelli-Paint}}.

\begin{figure}[t]
\centering
\includegraphics[width=0.48\textwidth]{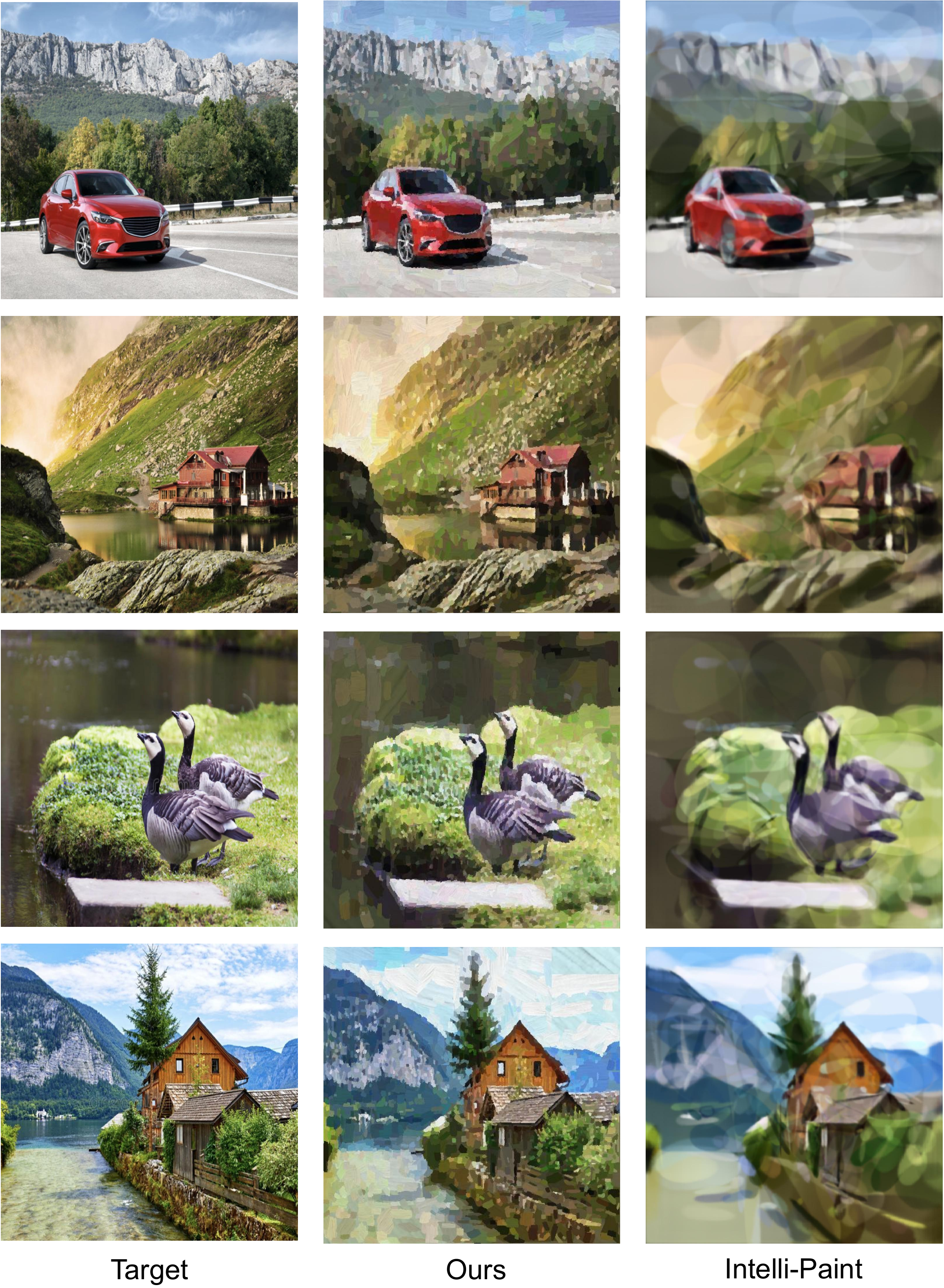}
\caption{The comparison between our results and the best-painted images shown in the paper of Intelli-Paint. Our model can reconstruct much more intricate details than Intelli-paint. 
}
\label{fig:compare with Intelli-Paint2}
%\vspace{-0.1in}
\end{figure}

In order to further validate our model's ability to capture more intricate details than Intelli-Paint without the need for object detection, we select several images from the best-performing examples in the paper of Intelli-Paint. We then employ our model to perform painting on these images and compared the results. As illustrated in Figure ~\ref{fig:compare with Intelli-Paint2}, our model achieved noticeably superior painting results, reconstructing much more intricate image details than Intelli-Paint.

\section{More Comparison on Stroke-based Painting}
\label{Sec:More Comparison on Stroke-based Painting}
\subsection{More comparison on oil brushstrokes}

In Section 4.3 of the main paper, we compare our model with the state-of-the-art stroke-based painting methods \cite{huang2019learning,liu2021paint,zou2021stylized,kotovenko2021rethinking,singh2021combining,tong2022im2oil}. In this section, to further validate our model's ability in stroke-based neural painting, we show more comparison results.

% In this section, we  compare with  the state-of-the-art SBR models and show more details on the boundary inconsistency problem of the existing methods \cite{huang2019learning,liu2021paint,zou2021stylized}.
% The comparison results are shown in figure \ref{fig:more comparison on image to paint}, \ref{fig:more comparison on boundary artifacts} and \ref{fig:more comparison on boundary artifacts2}. 

\begin{figure*}[t]
\centering
\includegraphics[width=0.95\textwidth]{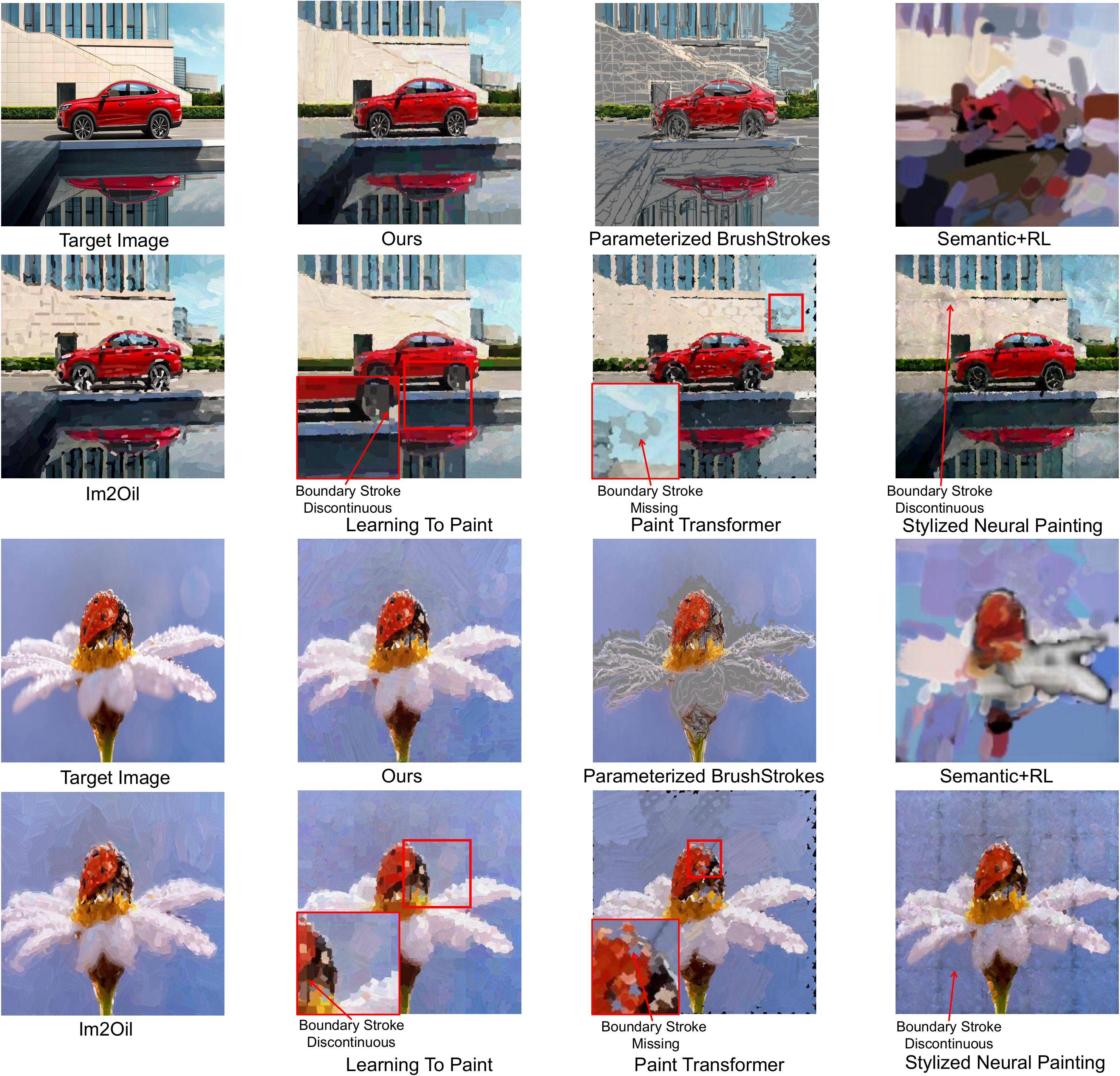}
%\vspace{-0.1in}
\caption{More comparison results to the existing methods (3,000 strokes). The methods which employ the uniform-block-dividing strategy (Learning To Paint, Paint Transformer and Stylized Neural Painting) all suffer from the boundary inconsistency artifacts. Please zoom in for details.
% (The Boundary discontinuous problem in Learning to Paint ) 
}
\label{fig:more comparison on image to paint}
%\vspace{-0.1in}
\end{figure*}

In Fig.~\ref{fig:more comparison on image to paint}, all methods use \hut{3,000} strokes to paint the images. Parameterized Brushstrokes~\cite{kotovenko2021rethinking} and Semantic + RL~\cite{singh2021combining} fail to reconstruct most of the details in the target image. Note that RL+Semantic requires the prediction of semantic region, and when the semantic region is wrongly predicted (the second and third examples in Fig.~\ref{fig:more comparison on image to paint}), its results lose most of the information in the target image.
In contrast, our method reconstructs the target images better under the same number of strokes.

\begin{figure*}[t]
\centering
\includegraphics[width=0.95\textwidth]{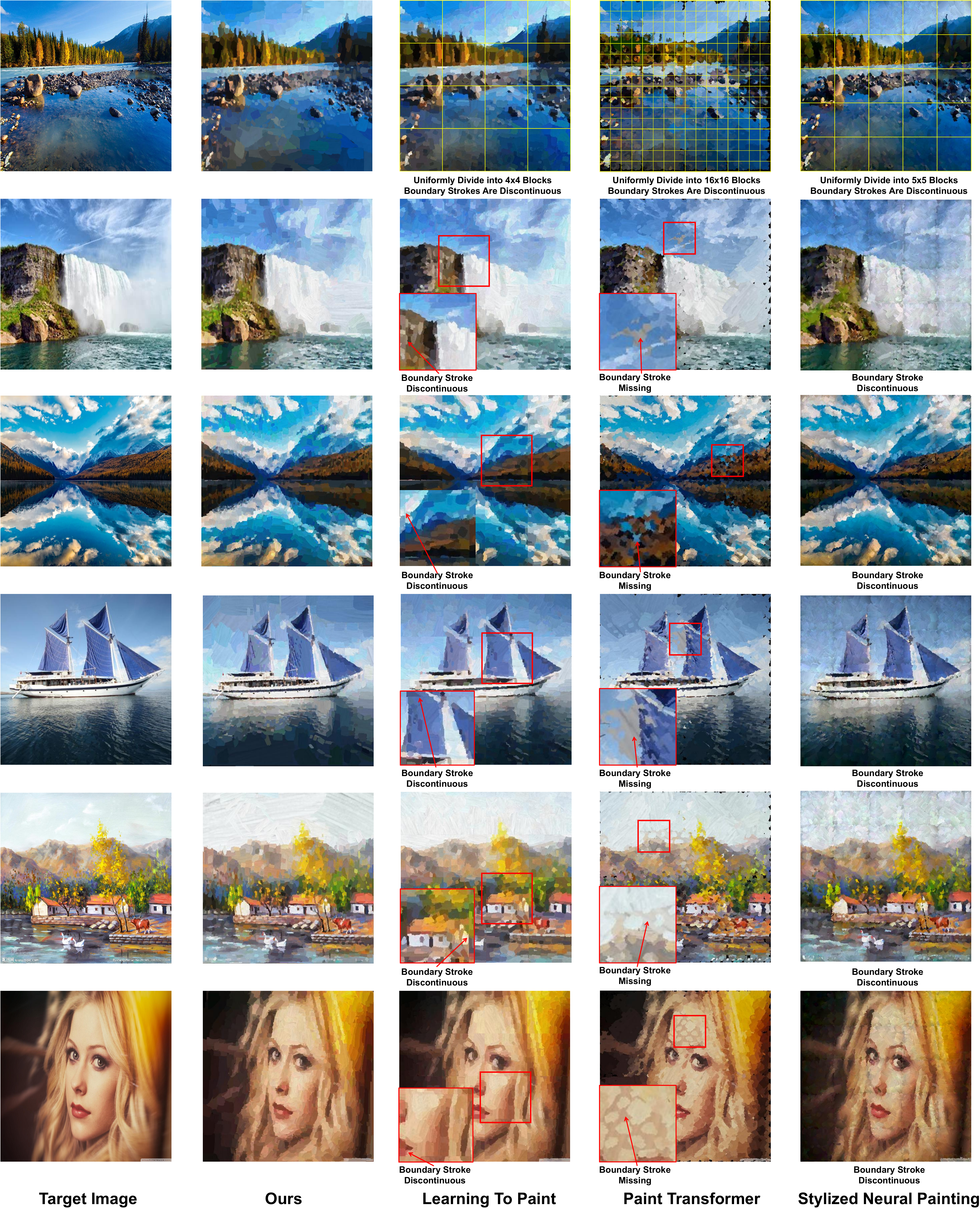}
\caption{More comparison results (5,000 strokes) with the methods which employ the uniform-block-dividing strategy. All the compared methods (Learning To Paint, Paint Transformer and Stylized Neural Painting) suffer from the boundary inconsistency artifacts while our model is free from the boundary inconsistency artifacts and has a better reconstruction quality. 
}
\label{fig:more comparison on boundary artifacts}
%\vspace{-0.1in}
\end{figure*}

\begin{figure*}[t]
\centering
\includegraphics[width=0.95\textwidth]{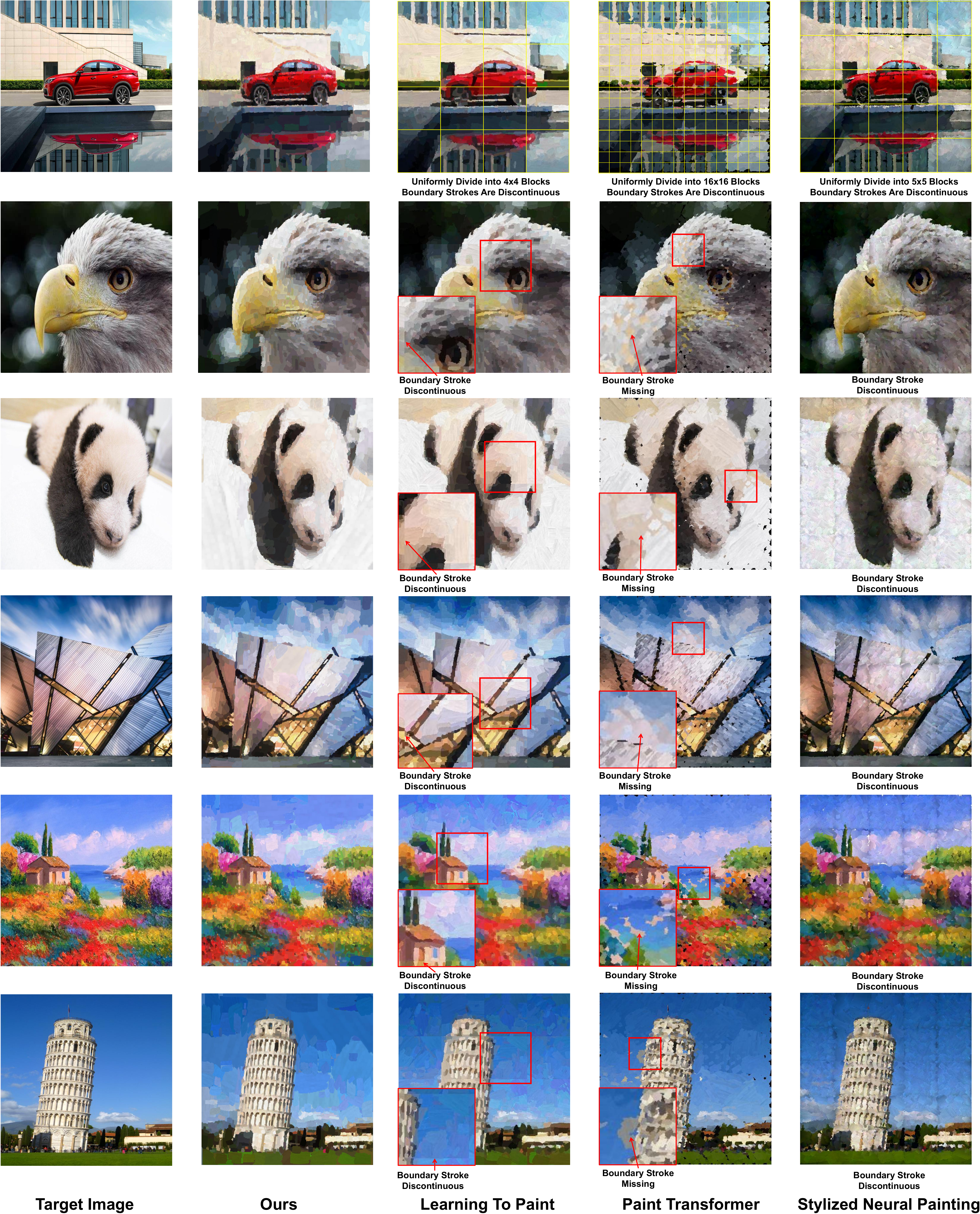}
\caption{More comparison results (5,000 strokes) with the methods which employ the uniform-block-dividing strategy. All the compared methods (Learning To Paint, Paint Transformer and Stylized Neural Painting) suffer from the boundary inconsistency artifacts while our model is free from the boundary inconsistency artifacts and has a better reconstruction quality. 
}
\label{fig:more comparison on boundary artifacts2}
%\vspace{-0.1in}
\end{figure*}

\begin{figure*}[t]
\centering
\includegraphics[width=0.95\textwidth]{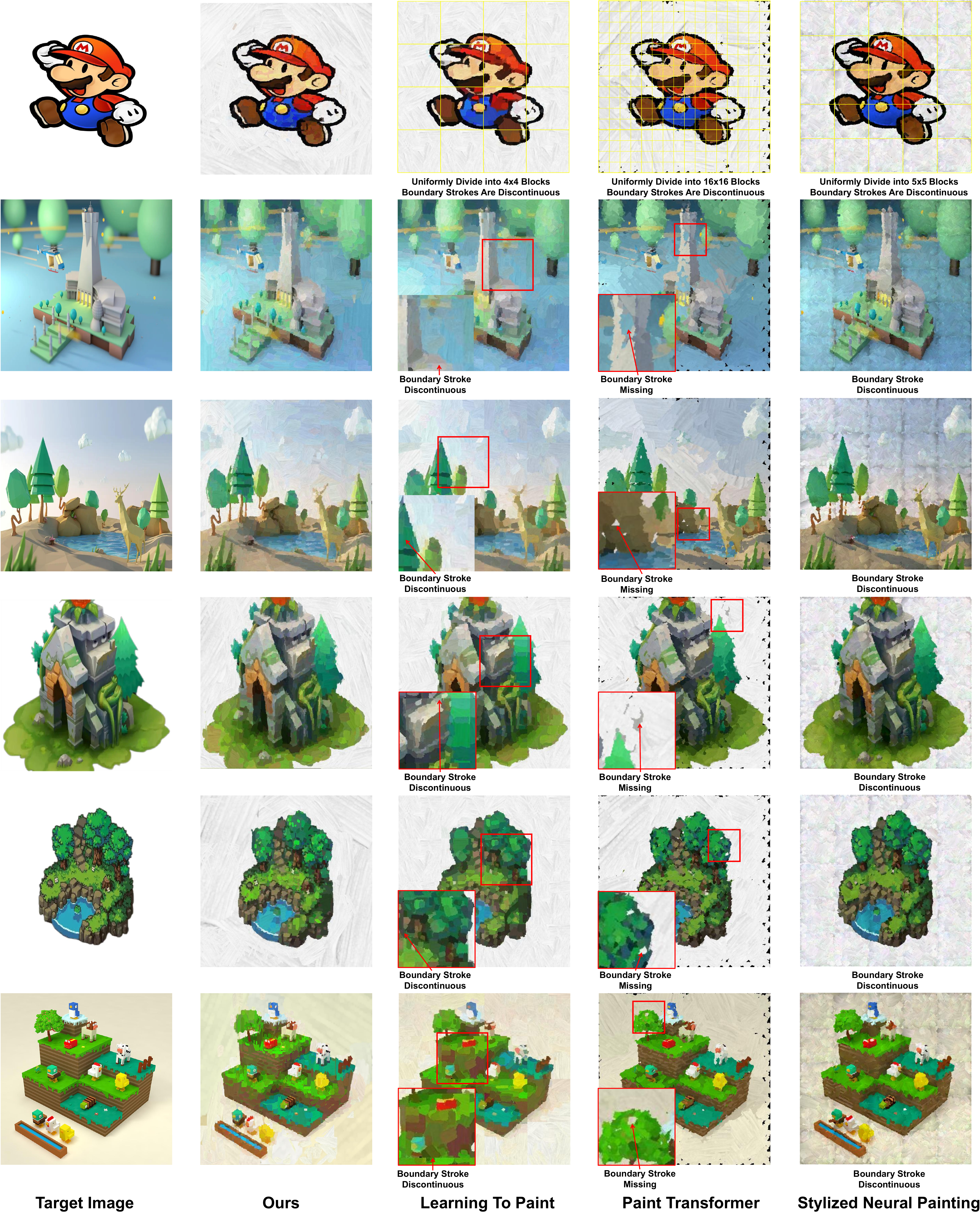}
\caption{More comparison results (5,000 strokes) with the methods which employ the uniform-block-dividing strategy. All the compared methods (Learning To Paint, Paint Transformer and Stylized Neural Painting) suffer from the boundary inconsistency artifacts while our model is free from the boundary inconsistency artifacts and has a better reconstruction quality. 
}
\label{fig:more comparison on boundary artifacts3}
%\vspace{-0.1in}
\end{figure*}

Learning To Paint~\cite{huang2019learning}, Paint Transformer~\cite{liu2021paint} and Stylized Neural Painting~\cite{zou2021stylized} are three comparison methods that utilize the uniform-block-dividing strategy, i.e., dividing the image plane uniformly into blocks, and then predicting strokes for each block independently, which enables a better reconstruction performance. 
However, they suffer from the boundary inconsistency artifacts: as shown in Fig.~\ref{fig:more comparison on image to paint} and more examples in Figs.~\ref{fig:more comparison on boundary artifacts}-\ref{fig:more comparison on boundary artifacts3} \hut{(5,000 strokes)}, there are obvious discontinuous strokes on the two sides of the block boundaries.
In contrast, with dynamically predicted painting regions, our model is free from the boundary inconsistency artifacts and paints the target images with the most details.

%We further show more comparison results with the methods using uniform-block-dividing strategy,
%To get a more solid conclusion that our model outperfomrs the existing state-of-the-art methods, we show more comparison results with the methods which can reconstruct the images in a relatively better way (i.e., the uniform-block-dividing methods: 
%i.e., Learning To Paint, Paint Transformer and Stylized Neural Painting, . 
%It can be seen that the methods with uniform-block-dividing strategy (dividing the image plane uniformly into blocks, and then predicting strokes for each block independently) suffer from boundary inconsistency artifacts (discontinuous strokes on the two side of the boundary). In contrast, our model paint the image much better than the state-of-the-art methods.

\subsection{Comparison with transparent brushstrokes}
\hut{In Learning To Paint~\cite{huang2019learning}, it employs transparent brushstrokes constructed by a large number of circles of different sizes to paint the images. It differs from real oil brushstrokes in terms of transparency and stroke texture. But to show the effectiveness of our method, we still compare with the state-of-the-art methods  (Learning To Paint~\cite{huang2019learning} and Semantic Guidance+RL\cite{singh2021combining}) which uses transparent brushstrokes in their default settings. In this experiment, we employ transparent burshstroke to train our model and compare it with the given model from their official links.  We use  $\mathcal{L}_2$ distance, PSNR and LPIPS as the comparison metric and paint 1,000 images in both ImageNet ~\cite{deng2009Imagenet} and CelebA-HQ\cite{karras2018progressive} dataset. The quantitative comparison results are shown in Tab.~\ref{tab:quantitative comparison on transparent brushstrokes}. It can be seen that our method outperforms both Learning To Paint and Semantic Guidance+RL with their transparent brushstrokes, indicating that our model has the same superior performance and strong robustness under different kinds of brushstrokes. 
Note that when painting images in CelebA-HQ dataset with 5,000 strokes, due to the simple structure and few details in human faces, our model performs similarly to Learning To Pain. But when the images become more complex (i.e., images in ImageNet), our model surpasses Learning To Paint in terms of both the reconstruction ability and perceptual similarity.}
\begin{table}[t]
\small
\centering
\setlength{\abovecaptionskip}{4pt}
\setlength{\belowcaptionskip}{-0.2cm}
\setlength\tabcolsep{3pt}
\renewcommand{\arraystretch}{1.2}
%\vspace{0.1in}
\caption{\hut{The quantitative comparison between the state-of-the-art methods and our model with the original transparent brushstrokes in Learning To Paint. 
% We conduct experiments on a 24G RTX3090 GPU, where Stylized Neural Painting \cite{zou2021stylized} can render at most 2,400 strokes due to excessive demand for memory, so there are no results for Stylized Neural Painting under 5,000 strokes.
}}
\scalebox{0.7}{
\begin{tabular}{cc|ccc|ccc}
\toprule
 Stroke&\multirow{2}{*}{Method}
 & \multicolumn{3}{c|}{ImageNet}& \multicolumn{3}{c}{Celeba-HQ}\\
 Num&&$\mathcal{L}_2$ Dist $\downarrow$&PSNR $\uparrow$&LPIPS $\downarrow$&$\mathcal{L}_2$ Dist $\downarrow$&PSNR $\uparrow$&LPIPS $\downarrow$\\
% Num                  &                            & $\mathcal{L}_2$ Dist $\downarrow$ & PSNR $\uparrow$ & LPIPS $\downarrow$ & $\mathcal{L}_2$ Dist $\downarrow$ & PSNR $\uparrow$ & LPIPS $\downarrow$ \\
\midrule
\multirow{3}{*}{200}
& Learning To Paint ~\cite{huang2019learning}
&\hut{0.0126}          & \hut{19.72}          & \hut{0.1658}          & \hut{0.0087}          & \hut{20.84}          & \hut{0.1534} \\
& Semantic Guidance+RL \cite{singh2021combining} &\hut{0.0126}&\hut{19.61}&\hut{0.2034}&\hut{0.0084}&\hut{20.97}&\hut{0.1829}  \\
& Ours                       & \textbf{\hut{0.0073}} & \textbf{\hut{22.07}} & \textbf{\hut{0.1593}} & \hut{\textbf{0.0039}} & \hut{\textbf{24.31}} & \hut{\textbf{0.1307}}\\        
\midrule

\multirow{3}{*}{500}
& Learning To Paint ~\cite{huang2019learning}  & \hut{0.0082} & \hut{21.81} & \hut{\textbf{0.1398}}  &  \hut{0.0032}  &  \hut{25.36}  &  \hut{0.1098}\\
& Semantic Guidance+RL \cite{singh2021combining}&\hut{0.0118}&\hut{19.90}&\hut{0.2019}&\hut{0.0076}&\hut{21.46}&\hut{0.1818}   \\
 & Ours                       & \textbf{\hut{0.0060}} & \textbf{\hut{23.29}} & \hut{0.1417} & \textbf{\hut{0.0025}} & \textbf{\hut{26.37}} & \hut{\textbf{0.1034}}     \\ 
\midrule

\multirow{3}{*}{1,000} 
& Learning To Paint ~\cite{huang2019learning} & \hut{0.0058} & \hut{23.49} & \hut{0.1179} &  \hut{0.0020}  &  \hut{27.44}  &  \hut{0.0853}\\
& Semantic Guidance+RL \cite{singh2021combining}&\hut{0.0117}&\hut{19.97}&\hut{0.2028}&\hut{0.0073}&\hut{21.65}&\hut{0.1786}   \\
& Ours                       & \textbf{\hut{0.0047}} & \textbf{\hut{24.59}} & \textbf{\hut{0.1154}} & \textbf{\hut{0.0018}} & \textbf{\hut{27.80}} & \textbf{\hut{0.0850}} \\
\midrule

\multirow{3}{*}{\hut{3,000}} 
& Learning To Paint ~\cite{huang2019learning} & \hut{0.0039} & \hut{25.31} & \hut{0.0975} &  \hut{\textbf{0.0013}}  &  \hut{29.23}  &  \hut{0.0670}\\
& Semantic Guidance+RL \cite{singh2021combining} &\hut{0.0117}&\hut{19.99}&\hut{0.2030}&\hut{0.0076}&\hut{21.50}&\hut{0.1802}  \\

& Ours                       & \textbf{\hut{0.0033}} & \textbf{\hut{25.99}} & \textbf{\hut{0.0956}} & \textbf{\hut{0.0013}} & \textbf{\hut{29.24}} & \textbf{\hut{0.0667}} \\
\midrule

\multirow{3}{*}{5,000} 
& Learning To Paint ~\cite{huang2019learning} & \hut{0.0036} & \hut{25.71} & \hut{0.0943} & \hut{\textbf{0.0012}}  &  \hut{\textbf{29.49}}  &  \hut{0.0658}\\

& Semantic Guidance+RL \cite{singh2021combining}&\hut{0.0116}&\hut{20.03}&\hut{0.2031}&\hut{0.0073}&\hut{21.69}&\hut{0.1791}   \\

& Ours                       & \textbf{\hut{0.0030}} & \textbf{\hut{26.35}} & \textbf{\hut{0.0905}} & \textbf{\hut{0.0012}} & \textbf{\hut{29.49}} & \textbf{\hut{0.0637}} \\ 
\bottomrule
\end{tabular}}
\label{tab:quantitative comparison on transparent brushstrokes}
%\vspace{-0.1in}
\end{table}

\section{More Stroke-based Stylization Experiments}
\label{Sec:More experiments on Stroke-based Stylization}
% \subsection{More Comparison Results %to the Existing Stylization Method
% }
In Section 4.4 of the main paper, we compare our model with the state-of-the-art stroke-based stylization methods: Stylized Neural Painting \cite{zou2021stylized} and Parameterized Brushstrokes \cite{kotovenko2021rethinking}. In this section, we take the stylization methods in pixel level into consideration and compare with them.

We compare our stroke-based stylization model with the existing methods, including stroke-based stylization methods: Stylized Neural Painting \cite{zou2021stylized} and Parameterized Brushstrokes \cite{kotovenko2021rethinking}, and pixel-based stylization methods: Gatys \cite{gatys2016image}, AdaIN  \cite{huang2017arbitrary} and AdaAttN \cite{liu2021adaattn}. 
%Note that we assign 2400 strokes (the maximum number of strokes which can be optimized by a single 24G GPU) for Stylized neural painting and more than 10000 strokes (the standard parameter in the source code) for Parameterized Strokes while our model only uses 2,000 strokes. 
\ori{We use the default stroke number for the stroke-based methods: 2,000 strokes for Stylized Neural Painting and Ours, and 10,000+ strokes for Parameterized Brushstrokes.}
We employ the official implementations from Github for all the comparison methods to conduct this experiment.

\begin{figure*}[t]
\centering
\includegraphics[width=0.95\textwidth]{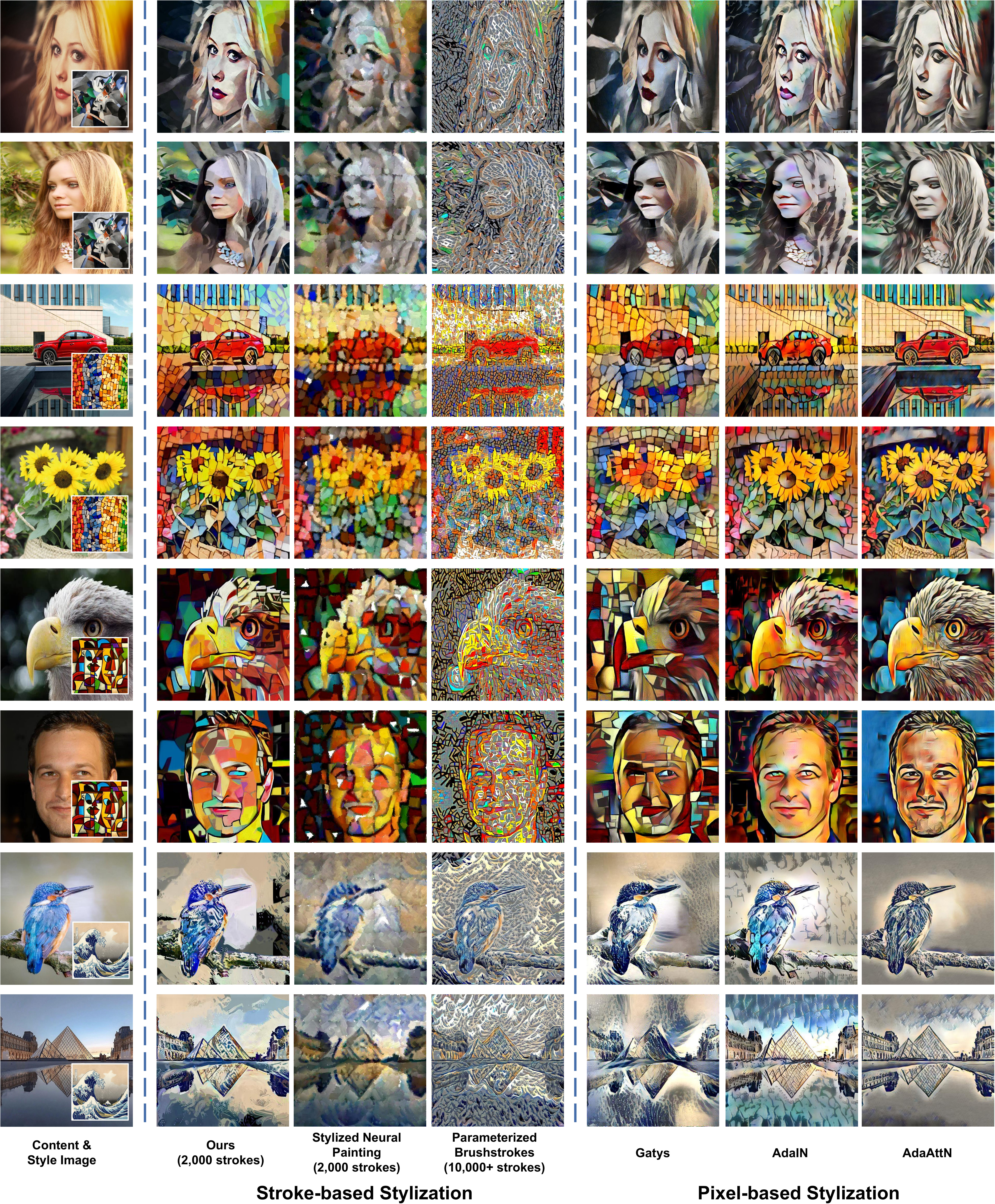}
%\vspace{-0.1in}
\caption{Stylization comparison with the existing methods. Compared to the stroke-based stylization methods (Stylized Neural Painting and Parameterized Brushstrokes), our model better preserves semantic contents in the content images and has a better visual quality. \ori{And our stroke-based model achieves stylization results as good as pixel-based methods (Gatys, AdaIN and AdaAttN).} %our model \ori{xxx} captures the styles better especially when the style images are composed of strokes (3rd to 6th rows).
}
\label{fig:more stylization comparison}
%\vspace{-0.1in}
\end{figure*}

The comparison results are shown in Fig.~\ref{fig:more stylization comparison}. 
Compared to the stroke-based stylization methods (Stylized Neural Painting and Parameterized Brushstrokes), our model better preserves semantic contents in the content images and has a better visual quality. 
%Compared to the pixel-based methods (Gatys, AdaIN and AdaAttn), our model captures the styles better, especially when the style images are composed of strokes (3rd to 6th rows).
\ori{And our stroke-based model achieves stylization results as good as pixel-based methods (Gatys, AdaIN and AdaAttN).}

\section{Time Complexity Analysis}
\label{Sec:time complexity}
\hut{To have a more comprehensive comparison, we further compare the inference time of each method on a single 24G RTX3090 GPU. Specifically, we set the stroke number to be 1,000 and we randomly select 100 images from the Celeba-HQ dataset to test the inference time. For each method, we use the default settings from the official code and only count the model's running time (i.e., ignore the loading and storage time). We average the inference time for the selected 100 images and report it in Tab.~\ref{tab:trainig and inference time}. We can see that Learning to Paint has the fastest inference speed (0.2066s) while our model follows closely behind (0.2162s), indicating that our model has both a good painting performance and efficiency.}

\begin{table}[t]
\small
\centering
\setlength{\abovecaptionskip}{2pt}
\setlength{\belowcaptionskip}{-0.2cm}
\setlength\tabcolsep{3pt}
\renewcommand{\arraystretch}{1.2}
\caption{\hut{Comparison on the average inference time.}}
\scalebox{0.95}{
\begin{tabular}{c|cccc}
\toprule
Model&Ours&\makecell[c]{Learning \\To Paint}&\makecell[c]{Semantic\\+RL}&\makecell[c]{Paint \\Transformer}\\\hline
Inference Time &\underline{0.2162s}&\textbf{0.2066s}&2.9921s&0.3725s\\\hline
Model&Im2Oil&\makecell[c]{Stylized Neural \\ Painting}&\makecell[c]{Parameterized \\Brushstrokes}&\\ \hline
Inference Time&55.3714s&124.9371s&247.6105s\\
\bottomrule
\end{tabular}}
\label{tab:trainig and inference time}
%\vspace{-0.05in}
\end{table}
% \subsection{Comparison with Pixel Loss}
% We first compare our DT loss with pixel loss in Figure \ref{fig:pixel loss vs DT loss} by optimizing the strokes' parameters.  we aims to cover the edge in target image with the stroke edges. When starting from a source stroke edge map which do not overlap with the target edge, pixel loss fails to cover the target edge while our DT loss can cover it well.
% \begin{figure}[t]
% \centering
% \includegraphics[width=0.45\textwidth]{figures/pixel loss vs DT loss.pdf}
% \caption{The comparison between pixel loss and DT loss in edge matching by parameter optimization. The pixel loss is not differentiable when the edges are not overlapped while our DT loss can be get a good performance.}
% \label{fig:pixel loss vs DT loss}
% %\vspace{-0.1in}
% \end{figure}

%\section{Paint With User Control}
%Our model can help users to participate in the painting process by utilizing the DT loss and CLIP \cite{radford2021learning}  loss. Specifically, we can

%\label{Sec:Paint With Use Control}

%\section{More Results of Our Method}
%In this section, we show more high-definition stroked-based paintings generated by our model.
%\label{More Results to Show}

\end{document}